\documentclass[
twocolumn,
tightenlines,
10pt,
longbibliography,
longtitle,
showpacs,
nofootinbib,
notitlepage,
superscriptaddress]{revtex4-2}

\usepackage{amsmath, amssymb, amsthm, braket, enumitem, dsfont, relsize}
\usepackage{makecell, graphicx, subfiles, multirow}
\usepackage[font=scriptsize]{subcaption}
\usepackage[font=scriptsize, justification=raggedright]{caption}
\usepackage[page]{appendix}
\usepackage[colorlinks, allcolors=black]{hyperref}
\usepackage[capitalize, nameinlink]{cleveref}
\usepackage{xcolor, tabularx, tikz}
\usetikzlibrary{quantikz}
\usetikzlibrary{shapes.geometric, arrows}
\usetikzlibrary{positioning}
\setlist[itemize]{leftmargin=10pt}
\usepackage[utf8]{inputenc}

\DeclareCaptionLabelFormat{bold}{\textbf{(#2)}}
\def\ptarget{{p}}
\def\KL{d_{\text{KL}}}
\DeclareMathOperator*{\argmax}{arg\,max}
\DeclareMathOperator{\plog}{polylog}
\renewcommand{\epsilon}{\varepsilon}
\renewcommand{\tilde}{\widetilde}
\renewcommand{\vec}[1]{\overrightarrow{#1}}

\mathcode`l="8000
\begingroup
\makeatletter
\lccode`\~=`\l
\DeclareMathSymbol{\lsb@l}{\mathalpha}{letters}{`l}
\lowercase{\gdef~{\ifnum\the\mathgroup=\m@ne \ell \else \lsb@l \fi}}%
\endgroup

\begin{document}

\title{Stochastic Security as a Performance Metric\\for Quantum-enhanced Generative AI}

\author{Noah~A.~Crum}
\email{ncrum@vols.utk.edu}
\affiliation{Department of Physics \& Astronomy, The University of Tennessee, Knoxville, TN 37996-1200, USA}

\author{Leanto~Sunny}
\email{lsunny@vols.utk.edu}
\affiliation{Department of Physics \& Astronomy, The University of Tennessee, Knoxville, TN 37996-1200, USA}

\author{Pooya~Ronagh}
\email{pooya.ronagh@uwaterloo.ca}
\affiliation{Institute for Quantum Computing, University of Waterloo, Waterloo, ON, N2L 3G1, Canada}
\affiliation{Department of Physics \& Astronomy, University of Waterloo, Waterloo, ON, N2L 3G1, Canada}
\affiliation{Perimeter Institute for Theoretical Physics, Waterloo, ON, N2L 2Y5, Canada}
\affiliation{1QB Information Technologies (1QBit), Vancouver, BC, V6E 4B1, Canada}

\author{Raymond~Laflamme}
\email{laflamme@uwaterloo.ca}
\affiliation{Institute for Quantum Computing, University of Waterloo, Waterloo, ON, N2L 3G1, Canada}
\affiliation{Department of Physics \& Astronomy, University of Waterloo, Waterloo, ON, N2L 3G1, Canada}
\affiliation{Perimeter Institute for Theoretical Physics, Waterloo, ON, N2L 2Y5, Canada}

\author{Radhakrishnan~Balu}
\email{rbalu@umd.edu}
\affiliation{Computer \& Information Sciences Directorate, Army Research Laboratory, Adelphi, MD 21005-5069, USA}
\affiliation{Department of Mathematics, University of Maryland, College Park, MD 20742-4015, USA\vspace{5pt}}

\author{George~Siopsis}
\email{siopsis@tennessee.edu}
\affiliation{Department of Physics \& Astronomy, The University of Tennessee, Knoxville, TN 37996-1200, USA}

\date{\today}

\begin{abstract}
Motivated by applications of quantum computers in Gibbs sampling from
continuous real-valued functions, we ask whether such algorithms can provide
practical advantages for machine learning models trained on classical data and
seek measures for quantifying such impacts. In this study, we focus on
deep energy-based models (EBM), as they require continuous-domain Gibbs
sampling both during training and inference. In lieu of fault-tolerant quantum
computers that can execute quantum Gibbs sampling algorithms, we use the
Monte Carlo simulation of diffusion processes as a classical alternative. More
specifically, we investigate whether long-run persistent chain Monte Carlo
simulation of Langevin dynamics improves the quality of the representations
achieved by EBMs. We consider a scheme in which the Monte Carlo simulation of a
diffusion, whose drift is given by the gradient of the energy function, is used
to improve the adversarial robustness and calibration score of an independent
classifier network. Our results show that increasing the computational budget
of Gibbs sampling in persistent contrastive divergence improves both the
calibration and adversarial robustness of the model, suggesting a prospective
avenue of quantum advantage for generative AI using future large-scale quantum
computers.

\vspace{10pt}
\end{abstract}

\maketitle

\section{Introduction}
\label{sec:intro}

Identifying practical evidence for quantum advantage in quantum machine
learning (QML) is a challenging task. Given the astonishing recent pace of
advancements in computer vision \citep{ramesh2022hierarchical, rombach2022high}
and language models \citep{brown2020language, openai2023gpt4}, it remains
unclear whether future quantum computers---capable of executing quantum
computations fault tolerantly---can
outperform classical artificial intelligence (AI) at
reasonable cost. In the absence of large-scale quantum computers that can
facilitate a direct assessment of any hypothesized revolutionary advances,
complexity-theoretic analysis appears to be the only rigorous fallback.
However, two main caveats arise when relying on the asymptotic time and
space complexity of an algorithm: (a) The complexity of an algorithm often
depends not only on the highlighted factors such as the dimensionality of a
problem, but also on additional factors (e.g., precision, Lipschitz constants,
and smoothness or convexity measures). (b) An algorithm with seemingly
excellent asymptotic complexity can still fail to offer practical
advantage, especially on data-driven problems. For example, while the famous
simplex algorithm for linear programming has an exponential worst-case
complexity, it is in practice preferred over polynomial-time alternatives
such as the interior-point method. The situation is even more complex for
quantum computing, since any quantum advantage must also be weighed against
additional costly resources for protecting the computation from
decoherence---such as the energy required for isolation, cooling, and control
in cryogenic settings---as well as the overhead of quantum error correction,
fault-tolerant protocols, and classical decoding of quantum error-correcting
codes.

Recently, \cite{motamedi2022gibbs} analyzed the complexity of solving the
Fokker--Planck (or diffusion) equation using quantum ODE (ordinary differential
equation) solvers, thereby obtaining a quantum Gibbs sampler
for continuous real-valued potentials. For Morse functions (i.e., those with
non-singular critical points), and at finite temperatures (the regime of
interest in generative AI tasks), this algorithm provides an exponential
speedup in the precision, Morseness, and Lipschitz constants of the potential,
while increasing the complexity dependence on the dimension of the problem
from $d^3$ to $d^7$. At first glance, the slowdown with respect to dimension
may seem like a major drawback. However, in this paper we show that boosting
the precision of a Gibbs sampler can be highly beneficial for enhancing the
robustness of the resulting generative AI. We therefore ask whether there are
suitable figures of merit for a generative AI model that can serve as
quantitative measures for such an assessment. We
focus on energy-based models (EBM) in this case study, since a recent
resurgence of them has provided state-of-the-art performance in generative
AI \citep{du2019implicit, grathwohl2019your, song2021train}. While it has been
hypothesized that EBMs offer improved representations of data in measures such
as robustness of the model to adversarial attacks and the calibration of its
predictions \cite{grathwohl2019your, hill2020stochastic}, both training and
inference from such models remain bottlenecked by the cost of Gibbs sampling
from an energy potential represented by the model network.

In this paper, we propose \emph{adversarial robustness} and
\emph{calibration scores} of EBMs trained with a Gibbs sampler (rather than
merely the sampling speed) as practical utility measures for assessing
the predicted quantum advantage.
Adversarial vulnerability is a critical obstacle to the reliable deployment
of autonomous agents in sensitive decision making scenarios.
\cite{du2019implicit} and \cite{grathwohl2019your} show that EBMs trained via
Langevin dynamics exhibit adversarial robustness without explicit adversarial
training. It is also important for such agents to be aware of the degree
of confidence in their decision, so that they can recognize scenarios in
which human expert intervention is required. Although
conventional supervised learning models have become more and more accurate in
their predictive capabilities, they have also become increasingly less
calibrated in their representations \citep{guo2017calibration}. In contrast,
\cite{grathwohl2019your} shows that EBMs can learn highly calibrated
representations of the CIFAR-10 and CIFAR-100 datasets. Other proposed
advantages of EBMs include improved mode coverage \citep{du2019implicit} and
better out-of-distribution (OOD) detection \citep{du2019implicit,
grathwohl2019your} compared to the autoregressive and flow-based generative
models.

To justify our choice, we numerically study the scaling of these utility
measures with respect to the number of iterations of stochastic gradient
Langevin dynamics (SGLD) during classical training of an EBM. We provide a
primer on EBMs in \cref{sec:EBMs}, and on the quantum Gibbs sampler
of \cite{motamedi2022gibbs} in \cref{sec:q-alg}.
In \cref{sec:training}, we use persistent contrastive divergence for classical
training of classical EBMs on the CIFAR-10 dataset.
Following \cite{hill2020stochastic}, we diffuse a test image via
the trained EBM's energy potential to purify it from
adversarial attacks, before passing it to a classifier network
trained completely independently from the EBM. As will
be shown, the same diffusion process can also result in more
calibrated logits for the classifier. We train a wide residual network
(WRN) 28-10 classifier \citep{Zagoruyko2016}, and multiple EBMs using
variable numbers of SGLD steps. Our results in
\cref{sec:adv-robustness,sec:calibration} show that purification of samples
using EBMs trained with more SGLD steps improves both the
adversarial robustness and the calibration of the WRN classifier. We visually
observe exponential decays in both the WRN's adversarial vulnerability and
its calibration error, and quantify these trends using linear regression. Our
results provide positive evidence for the utility of efficient Gibbs samplers
on continuous potentials, underscoring that the computational bottleneck of
EBMs is well worth overcoming using quantum computation.

\section{Energy-based models}
\label{sec:EBMs}

Conventional EBMs, such as Boltzmann machines and Hopfield neural
networks \citep{ackley1985learning, hopfield1982neural}, and their derivative
models \citep{hinton2006fast, hinton2007learning}, are built via undirected
graphical models \citep{koller2009probabilistic, clifford1971markov}. The
neurons represent discrete (binary) random variables and the model's energy
is a function of these discrete random variables (for example, a
quadratic binary-valued function representing two-body interactions between the
neurons). However, modern EBMs instead use deep neural networks as their
underlying graphical models. The deep neural network parametrizes a highly
non-convex energy potential defined on a continuous domain (i.e., the
continuous signals fed into network's input layer).

Such EBMs are trained using Monte Carlo integration of Langevin dynamics
\citep{du2019implicit, nijkamp2020anatomy}---a stochastic differential equation
(SDE) governing diffusion processes. Alternatively, corrections using the
Metropolis--Hastings criteria \citep{che2020your}, and second-order variations,
such as the Hamiltonian Monte Carlo can be applied \citep{chen2014stochastic}.
Regardless, these methods for Gibbs sampling require very high numbers
of Langevin steps (i.e., the Monte Carlo iterations on the SDE) to approach
mixing. Consequently, they are extremely costly and can be numerically unstable
in practice. Indeed, many prior works adopt non-convergent shortcuts to training
EBMs by collecting short-run Langevin samples \citep{nijkamp2019learning}. The
number of Langevin steps required for training truly convergent EBMs is
expected to be at least in the tens of thousands, even for low-resolution image
data \citep{nijkamp2020anatomy}.

An EBM consists of a graphical model that acts as a function approximator for
an \emph{energy potential} $E_\theta: \Omega \to \mathbb R$. Here
$\theta \in \mathbb R^D$ is a vector of $D$ model parameters,
$\Omega \subseteq R^d$ is a probability space endowed with a probability
measure, and $d$ represents the dimension of input signals. In this
paper we focus on image data without using latent diffusion
\citep{rombach2022high}. Therefore, $d$ is the number of pixels of the
images multiplied by its number of color channels. We also assume pixel
intensities are normalized between 0 and 1, so $\Omega = [0, 1]^d$
throughout. In state-of-the-art diffusion models, high-resolution images are
encoded and decoded into much smaller latent spaces, typically with only a few
hundred dimensions.

Similar to other ML models, this graphical model is used to determine a \emph
{model distribution}. Given a set of i.i.d training samples $\mathcal D=
\{x_1, \dots, x_N\} \subset \Omega$ the goal is to learn
a vector of model parameters $\theta^* \in \mathbb R^D$ such that the
resulting model distribution best approximates an unknown distribution
$\ptarget$ from which $\mathcal D$ is assumed to have been sampled. What
distinguishes EBMs is that the model distributions are Gibbs distribution
\begin{equation}
\label{eq:Gibbs-distribution}
p_\theta (x) = \exp(-\beta E_\theta (x))/ Z_{\beta, \theta}
\end{equation}
of the energy potential. The normalizing constant
$Z_{\beta, \theta} = \int_{x \in \Omega} \exp(-\beta E_\theta(x))\, dx$ is the
partition function of $p_\theta$ at thermodynamic constant $\beta$.
Since the model only determines the energy potential $E_\theta$ (and not
$p_\theta$ directly) it is called an \emph{unnormalized} probabilistic
graphical model. Unlike typical ML models used in discriminative tasks (e.g.,
classification), EBMs impose no constraints to ensure the tractability of the
normalization constant. This makes EBMs more expressive, although it comes at
the cost of computational intractability of both training and inference.

The standard method for training probabilistic models from i.i.d. data is
maximum likelihood training; i.e., maximizing the expected log-likelihood
function over the data distribution,
$\mathbb E_{x \sim \ptarget} [\log p_\theta(x)]$.
This is equivalent to minimizing the KL distance between $p_\theta$
and $\ptarget$, because
\begin{equation}
\label{eq:data-model-kl}
\KL(\ptarget(x)||p_\theta(x))=
-\mathbb E_{x \sim \ptarget} [\log p_\theta(x)] + \text{constant}.
\end{equation}
In fact, we do not require direct access to the likelihood itself but
only the gradient of the log-probability of the model. This holds at
least for first-order optimization schemes which are the standard techniques
for training classical ML models. We have
\begin{equation}
\label{eq:grad-log-prob}
\begin{split}
\nabla_\theta \log p_\theta(x)
&= -\beta \nabla_\theta E_\theta(x)
- \nabla_\theta \log Z_{\theta, \beta}\\
&= -\beta \nabla_\theta E_\theta(x)
+ \beta \mathbb E_{x \sim p_\theta}[\nabla_\theta E_\theta(x)].
\end{split}
\end{equation}

In EBMs, the first term above is easily calculated via automatic
differentiation, whereas the second term must be approximated through costly
Gibbs sampling. Indeed, if we can sample from the model distribution
$p_\theta$, we obtain unbiased estimates of $\nabla_\theta E_\theta(x)$,
which we can then leverage to train the EBM via stochastic gradient
descent. The overdamped Langevin diffusion
\begin{equation}
\label{eq:langevin}
dx= -\alpha \nabla_x E_\theta (x)\, dt + \sigma\, dW
\end{equation}
where $W$ is a standard Wiener process, is known to mix into our desired Gibbs
distribution at inverse temperature $\beta= 2 \alpha / \sigma^2$. Monte Carlo
simulation of this SDE is known as stochastic
gradient Langevin dynamics (SGLD) and amounts to building discrete time chains
\begin{equation}
\label{eq:sgld}
\begin{split}
x_0 &\sim p_0 (x),\\
x_{i+1} &= x_i -\alpha \nabla E_\theta (x_i)\, dt
+ \epsilon_i, \quad \epsilon_i \sim N(0, \sigma),
\end{split}
\end{equation}
after a finite number of iterations, where the initial distribution $p_0$ is
typically a uniform distribution.

\section{The quantum Gibbs sampler}
\label{sec:q-alg}

\begin{figure}[t]
\centering
\begin{quantikz}
\lstick{$\ket{\theta}$} & \qw & \gate[3]{O_E} & \qw & \qw
\rstick{$\ket{\theta}$}\\
\lstick{$\ket{x}$} & \qw & & \qw & \qw \rstick{$\ket{x}$} \\
\lstick{$\ket{y}$} & \qw & & \qw & \qw \rstick{$\ket{E_\theta(x) \oplus y}$}
\end{quantikz}
\caption{The oracle for the energy potential. All registers receive float-point
representations of real numbers. The first register receives the current model
parameters $\theta \in \mathbb R^D$. The second register receives a data
sample $x \in \Omega \subseteq \mathbb R^d$. And, the third register is used
to evaluate the energy potential.}
\label{fig:EBM-circuit}
\end{figure}

Simulating equilibrium dynamics has long been proposed as a potential
application of quantum computers for the past two decades
\citep{terhal2000problem,
poulin2009sampling, temme2011quantum, kastoryano2016quantum,
chowdhury2016quantum, van2020quantum, lemieux2020efficient,
bravyi2021complexity}. However, these references focus on Gibbs sampling from
discrete spin systems, thereby limiting their applicability to
traditional EBMs \citep
{wiebe2014quantum, amin2018quantum, crawford2016reinforcement, levit2017free,
sepehry2022quantum}. Fortunately, more recent works investigate quantum
advantage in Gibbs sampling over continuous domains
\citep{childs2022quantum, motamedi2022gibbs}. \cite{childs2022quantum}
considers replacing the classical Monte Carlo simulation of Langevin dynamics
with quantum random walks. Consequently, their analysis is limited to convex
potentials, which are not relevant to EBMs. However,
\cite{motamedi2022gibbs} solves the Fokker--Planck equation (FPE)
and prove quantum speedups for highly non-convex potentials. Nevertheless,
both of these quantum algorithms require large-scale fault-tolerant quantum
computers to train a model with as many parameters as a modest-size modern EBM.
Therefore, it is crucial to ask whether a prospective quantum acceleration of
continuous-domain Gibbs sampling can serve as a practical motivation for
building large fault-tolerant quantum computers.

A quantum Gibbs sampler can replace SGLD as described in \cref{sec:EBMs} both
during the training of and inference from an EBM. The neural network of
the EBM is a (white-box) composition of affine transformations with nonlinear
activation functions. In gate-based quantum computing, one can efficiently
construct quantum circuits
implementing the arithmetic required for realizing each such function
with only a polylogarithmic overhead compared to the gate complexity of the
corresponding classical boolean circuit \citep{nielsen2002quantum}. We can
therefore consider a quantum circuit counterpart for the classical deep
neural network endowed with a register for input samples $\ket{x}$ in the
computational basis, and another register for receiving the model parameters
$\ket{\theta}$. Alternatively, the model parameters may be encoded at
compile time in a QROM to reduce the qubit count of the circuit
\citep{babbush2018encoding}.
\cref{fig:EBM-circuit} provides a schematic representation of such
a quantum circuit, and will be queried as an \emph{oracle} by the Gibbs
sampler. More specifically, $O(d^2)$ replicas of this circuit are used to
construct a controlled unitary operation which is executed repeatedly by the
quantum Gibbs sampler. It is important to highlight that unlike most quantum
algorithms proposed for classical-data problems, this scheme does not rely on
preparing a superposition of classical data in a QRAM, which is extremely
costly \citep{arunachalam2015robustness, di2020fault}. Instead, the classical
training data is provided iteratively in the computational basis to the $\ket
{x}$ register, similar to how data is used in classical ML.

For our experiments in \cref{sec:training} we train the EBM architecture
of \cite{nijkamp2019learning}, shown in \cref{fig:EBM-model}. Since we do not
encode the image in a latent space, the domain of the energy potential has the
fairly large dimension of 3072. However, even such a large domain is not beyond
the practical reach of QROMs. An educated guess of code distances required with
emerging superconducting qubits suggests that code distances below $\sim50$
suffice for very large computations \citep{mohseni2024build}. This
gives us an estimate of $\sim2.5$ million qubits required for a QROM acting on
a 3072-wide register of 16-bit float-point numbers. Therefore, we expect these
quantum oracles will also be feasible to implement on utility-scale quantum
computers. However, we leave a careful and end-to-end resource estimation on
our algorithm for future efforts. \cite{motamedi2022gibbs} shows that expanding
the FPE in the Fourier basis and solving it via a quantum ODE solver yields
high-precision solutions for highly non-convex (but periodic)
functions. We now provide a brief and informal summary of the steps of this
algorithm.

\begin{figure}[t]
\centering
\resizebox{0.9\columnwidth}{!}{
\begin{tikzpicture}[
    font=\sffamily,
    node distance=0.5cm,
    >={Stealth[length=2mm]},
    line width=.5pt,
    io/.style={
        rectangle,
        rounded corners=3pt,
        draw=black,
        fill=gray!40,
        align=center,
        minimum width=2.5cm,
        minimum height=1cm
    },
    conv/.style={
        rectangle,
        rounded corners=3pt,
        draw=black,
        fill=cyan!60!black,
        text=white,
        align=center,
        minimum width=3cm,
        minimum height=1.2cm
    },
    leakyrelu/.style={
        rectangle,
        rounded corners=3pt,
        draw=black,
        fill=red!60!black,
        text=white,
        align=center,
        minimum width=2cm,
        minimum height=0.75cm
    },
    squeeze/.style={
        rectangle,
        rounded corners=3pt,
        draw=black,
        fill=teal!100!black,
        text=white,
        align=center,
        minimum width=2.5cm,
        minimum height=1cm
    },
    breaknode/.style={
        text width=1cm,
        align=center,
        font=\large
    },
    arr/.style={
        thick,->,
    }
]

\node[io] (input) {input tensor};

\node[below=5pt of input, font=\footnotesize] (dim0)
{\scriptsize $\mspace{130mu} 1 \times 3 \times 32 \times 32$};


\node[conv, below=25pt of input] (conv1) {
  \textbf{Conv}\\[3pt]
  \scriptsize W $\langle32 \times 3 \times 3 \times 3\rangle$\\
  B $\langle 32 \rangle$
};
\draw[arr] (input) -- (conv1);

\node[leakyrelu, below=of conv1] (lrelu1) {LeakyRelu};
\draw[arr] (conv1) -- (lrelu1);

\node[conv, below=of lrelu1] (conv2) {
  \textbf{Conv}\\[3pt]
  \scriptsize W $\langle64 \times 32 \times 4 \times 4\rangle$\\
  B $\langle 64 \rangle$
};
\draw[arr] (lrelu1) -- (conv2);

\node[leakyrelu, below=of conv2] (lrelu2) {LeakyRelu};
\draw[arr] (conv2) -- (lrelu2);

\node[conv, below=of lrelu2] (conv3) {
  \textbf{Conv}\\[3pt]
  \scriptsize W $\langle128 \times 64 \times 4 \times 4\rangle$\\
  B $\langle 128 \rangle$
};
\draw[arr] (lrelu2) -- (conv3);

\node[breaknode, below= of conv3] (downbreak_anchor) {}; 
\node[breaknode, yshift=-1pt] at (downbreak_anchor) (downbreak) {\vdots}; 
\draw[arr] (conv3) -- (downbreak_anchor);

\node[breaknode, right=3.5cm of input] (upbreak) {\vdots};

\node[leakyrelu, below=0.5cm of upbreak] (lrelu3) {LeakyReLU};
\draw[arr] (upbreak) -- (lrelu3);

\node[conv, below=of lrelu3] (conv4) {
  \textbf{Conv}\\[3pt]
  \scriptsize W $\langle256 \times 128 \times 4 \times 4\rangle$\\
  B $\langle 256 \rangle$
};
\draw[arr] (lrelu3) -- (conv4);

\node[leakyrelu, below=of conv4] (lrelu4) {LeakyRelu};
\draw[arr] (conv4) -- (lrelu4);

\node[conv, below=of lrelu4] (conv5) {
  \textbf{Conv}\\[3pt]
  \scriptsize W $\langle1 \times 256 \times 4 \times 4\rangle$\\
  B $\langle 1 \rangle$
};
\draw[arr] (lrelu4) -- (conv5);

\node[squeeze, below=of conv5] (sqz) {Squeeze};
\draw[arr] (conv5) -- (sqz);

\node[io, below=of sqz] (out) {energy value};
\draw[arr] (sqz) -- (out);

\end{tikzpicture}}
\caption{The EBM architecture of \cite{nijkamp2019learning}.
The neural network parameterizes a real-valued energy function which
defines the energy potential used by the quantum Gibbs sampler.}
\label{fig:EBM-model}
\end{figure}

The Fokker--Planck equation is the PDE (partial
differential equation) counterpart to the Langevin SDE of \cref{eq:langevin}
via the Feynamn-Kac formula \citep{pavliotis2014stochastic}. While the SDE in
\cref{eq:langevin} describes the stochastic transitions of a multi-dimensional
random variable $X_t$ attaining values $x \in \Omega$ as a function of time,
the FPE describes the time evolution of the entire probability density
function $\rho_t (x)$ on $\Omega$ from which $X_t$ is sampled. The FPE can be
written in terms of the second-order differential operator
$\mathcal{L}(-) = \nabla \cdot \left(
e^{-E} \,\nabla \left( e^{E} \, - \right) \right)$ as its generator:
\begin{align}
\label{eq:toroidal-FP}
\frac{d}{dt} \rho_t (x) &= \mathcal L \rho_t (x).
\end{align}

To integrate this equation for time $T>0$ with precision $\epsilon >0$, we
discretize the domain $\Omega$ as a regular lattice $\mathcal V_N$ with
$N= O(d\plog(\epsilon))$ equidistant points along each dimension, and formally
replace all differential operators with Fourier derivatives
(see Appendix~A of \cite{motamedi2022gibbs}) to obtain the linear operator
\begin{equation}\label{eq:def-dis-L}
\begin{split}
\mathbb L: \mathcal V_N &\to \mathcal V_N \\
\vec u &\mapsto \tilde{\nabla}
\cdot\left( e^{-E} \tilde{\nabla}(e^E \vec u)\right)
\end{split}
\end{equation}
where tilde signs over the derivatives represent Fourier differentiation,
and the arrow over $u$ indicates that the function is viewed
as the long vector of the values it attains on the discrete lattice points.

At an inverse temperature $\beta>0$, a total of
$O(\lambda^{-2} e^{\beta/2} d^7 \plog(\epsilon))$ queries to the energy
oracle are required to obtain $\epsilon$-accurate samples from the Gibbs
distribution of the energy potential $E$, if it is $\lambda$-strongly Morse
and periodic. These technical conditions are quite tame for machine learning
applications since a non-Morse function can be regularized into a Morse one
using regularizers and low temperatures $\beta \to \infty$ are not
desired in generative modeling as they reduce the generalization power of
the model. Finally, datasets from non-periodic patterns can be transformed into
periodic ones by inverting a trigonometric function of each component, as shown
in \cref{fig:arccos-EBM}. Here, each dimension of data is lifted to two branches
of the $\arccos$ function. Therefore, each sample $\ket{x}$ in the computational
basis corresponds to $2^d$ inverse images, and an equal superposition $\ket
{\Psi_x}$ of these exponentially many inverse images is efficiently constructed
using Hadamard gates. By employing this approach, the resulting energy function
learns a consistent periodic potential across all branches of $\arccos$.

\begin{figure}[b]
\centering
\begin{quantikz}
\lstick{$\ket{\theta}$} &\qw &\qw & \gate[3]{O_{E_\theta}} &\qw &\qw  \rstick{$\ket{\theta}$} \\
\lstick{$\ket{x}$} & \gate{\arccos} &\gate{H^{\otimes n}} &  & \qw & \qw\rstick{$\ket{\Psi_x}$} \\
\lstick{$\ket{y}$} & \qw &\qw & & \qw & \qw \rstick{$\ket{E_\theta(x) \oplus y}$}
\end{quantikz}
\caption{The oracle of \cref{fig:EBM-circuit} receiving
an augmented sample in superposition as the state
$\ket{\Psi_x}= \sum_{b \in \{0, 1\}^d}\ket{(-1)^{b_i} \arccos x_i}$.}
\label{fig:arccos-EBM}
\end{figure}

\section{Training with Persistent Chain Monte Carlo}
\label{sec:training}

\begin{figure}[t]
\centering
\includegraphics[width=0.9\columnwidth, trim=15mm 30mm 30mm 50mm, clip]{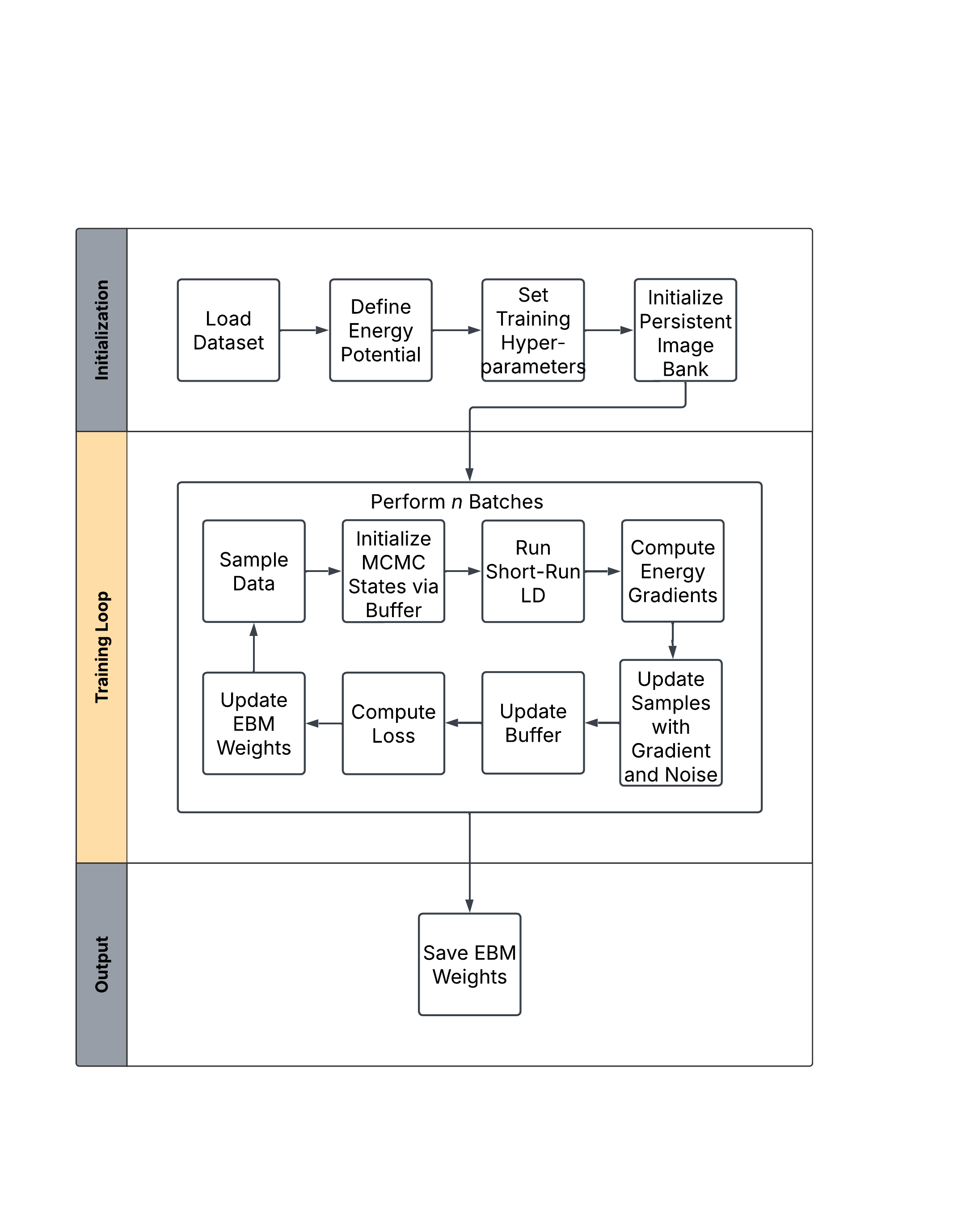}
\caption{Training of an EBM using persistent contrastive divergence (PCD)
applied to stochastic gradient Langevin dynamics (SGLD). The use of a replay
buffer that retains samples from older shorter runs of Langevin dynamics
significantly expedites the training process at the cost of creating
some instabilities as shown in \cref{eq:cd-gradient}.}
\label{fig:EBM-training}
\end{figure}

We train several EBMs on the CIFAR-10 dataset
\citep{Krizhevsky09learningmultiple} using SGLD as described in \cref{sec:EBMs}.
The training procedure uses persistent contrastive divergence (PCD)
\citep{tieleman2009using}, as shown in \cref{fig:EBM-training}.
Here, a replay buffer retains samples
from older, shorter runs of Langevin dynamics. This persistent-chain buffer
offers orders-of-magnitude savings in the simulation time of stochastic
Langevin dynamics---possibly more than thousands even for small
datasets~\citep{nijkamp2020anatomy}.
However, it comes with the cost of increased training instability.
Indeed, the distribution, $\pi$, representing the persistent chain differs
from the true model distribution $p_\theta$ (the Gibbs distribution).
Hence, the update direction followed by $\theta$ deviates from that of
\cref{eq:data-model-kl} and must instead be viewed as
\begin{equation}
\label{eq:cd-gradient}
\KL(\ptarget(x)||p_\theta(x)) - \KL(\pi||p_\theta(x)).
\end{equation}
Note that the second divergence is maximized during training. Therefore, the
persistent chain distribution diverges from the instantaneous Gibbs state during
training, and the faster the model parameters are updated (i.e., with higher
learning rate), this deviation increases.

We empirically observed that it is not quite feasible to train such EBMs
without PCD. Since we do not reinitialize the persistent chain samples,
each PCD sample undergoes all SGLD steps during training, which is
proportional to the number of SGLD steps in each epoch of training.
Despite the discrepancy in \cref{eq:cd-gradient}, this value can still be a
good indicator of the computational cost of Gibbs sampling for training EBMs.

We train five distinct EBMs with the architecture of \cref{fig:EBM-model}, each
with a different number of Langevin steps during PCD. Each EBM is trained for
250,000 batches, beginning with the Adam optimizer \citep{kingma2014method} and
switching to SGD after 125,000 batches with learning rates of 1e-4 and 5e-5
respectively. The change from Adam to SGD has been found to be essential for
stable and convergent training of EBMs \citep{nijkamp2019learning}. Details of
the hyperparameters used to train the models are listed in
\cref{tab:hyperparameters}.

\begin{table}[b]
    \centering
    \begin{tabular}{c|c|c}
     & WRN classifier & EBMs \\
    \hline
        Training duration & 100 epochs & 250,000 batches \\
        Batch size & 100 & 100\\
        Learning rates (LR)& [1e-1, 1e-2, 1e-3] & [1e-4, 5e-5] \\
        LR switch time & Epochs 40 and 60 & Batch 125,000 \\
        $L_2$ regularizer & 2e-4 & 0 \\
        SGLD steps, $n$ & - & [50,\! 75,\! 100,\! 150,\! 200] \\
        SGLD step size, $\alpha$ & - & 0.01 \\
        SGLD noise, $\sigma$ & - & 0.01 \\
    \end{tabular}
    \caption{Training hyperparameters. The first column pertains to training
    of the WRN classifier. The second column reflects the hyperparameters used
    for training EBMs with varied numbers of SGLD steps.}
    \label{tab:hyperparameters}
\end{table}

\section{Adversarial robustness}
\label{sec:adv-robustness}

Following \cite{hill2020stochastic}, we train a prototypical wide residual
network (WRN) 28-10 classifier \citep{Zagoruyko2016}. The classifier is trained
for 100 epochs with a scheduled learning-rate decay via SGD on a cross-entropy
loss function, with an $L_2$ regularizer of coefficient 2e-4 for the weights and
biases of the model. We denote the
classification labels by $y \in \mathcal Y= \{1, \ldots, 10\}$, and for an input
image $x$ we denote the output logits of the classifier by $f(x)$. We perform
white-box projected gradient descent (PGD) attacks on the
classifier \citep{madry2017towards}. Then we pass the attacked image
through Langevin dynamics of the EBMs trained in \cref{sec:EBMs} to purify the
image from the attack. We refer the reader to \cite[Appendix C]
{hill2020stochastic} for some theoretical justifications on this method based
on the memoryless and chaotic behavior of Markov chain Monte Carlo sampling.

\begin{figure}[b]
\centering
\includegraphics[width=0.9\columnwidth, trim=5mm 0mm 0mm 10mm, clip]{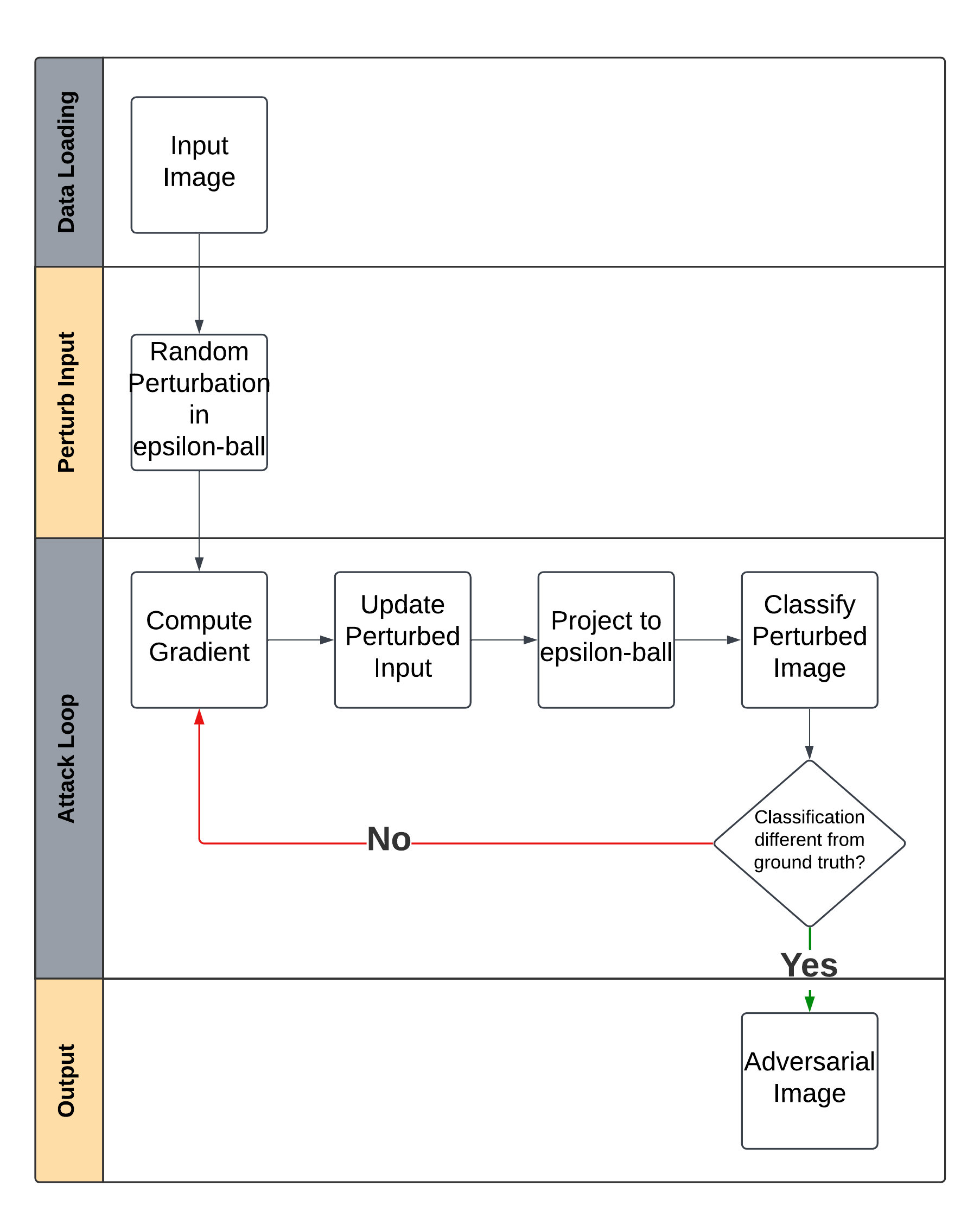}
\caption{The projected gradient descent (PGD) attack scheme of
\cite{madry2017towards}. We perform white-box PGD attacks  on the classifier.
We then pass the attacked image through Langevin dynamics of EBMs to purify it
from the attack.}
\label{fig:PGD-attack}
\end{figure}

\begin{figure}[t]
\centering
\includegraphics[width=0.9\columnwidth, trim=75mm 15mm 80mm 10mm, clip]{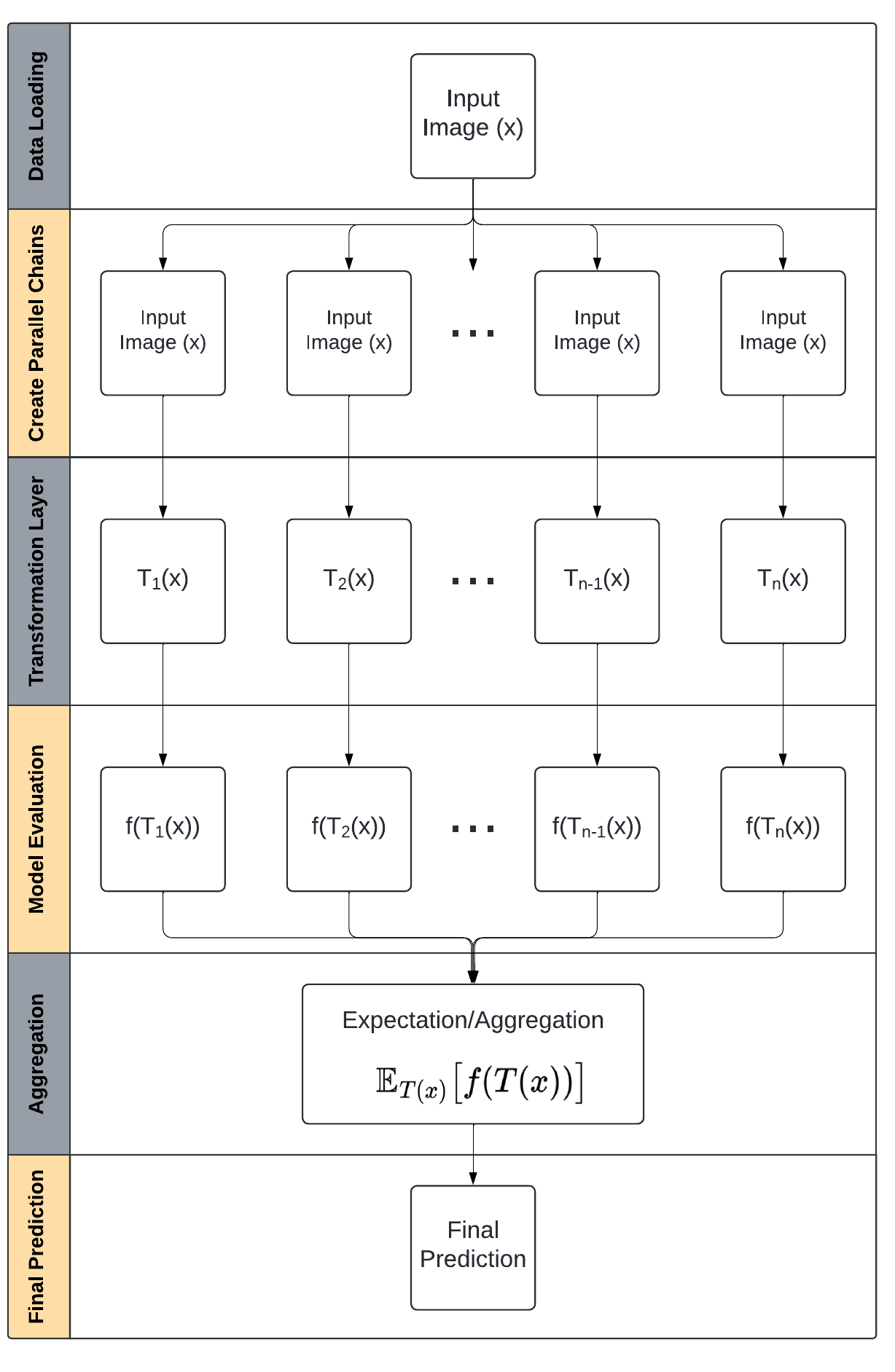}
\caption{The schematic of an expectation over transformation (EOT) defense.
Attacked images are diffused via an EBM in parallel trials. The outputs are
then provided to the classifier and the logits of all the trials are
averaged to provide a final post-transformation label.}
\label{fig:EOT}
\end{figure}

\begin{figure}[t]
\centering
\includegraphics[width=\columnwidth]{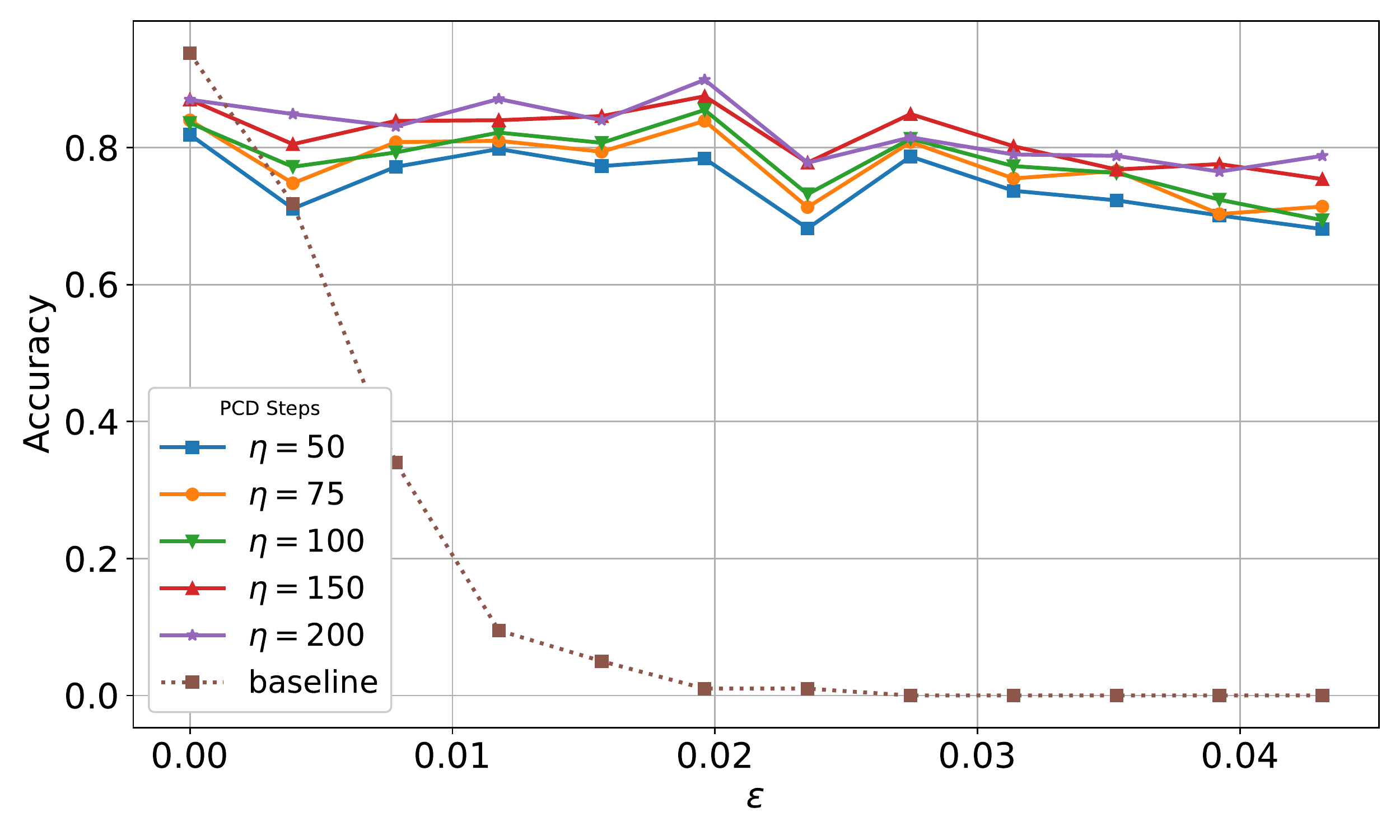}
\caption{Classification accuracy versus adversarial attack strength $\epsilon$.
 The `baseline' refers to the classifier unequipped with an EOT defense. The
 other curves represent the accuracy of the distinct EBMs at classification of
 1000 PGD adversarial images after purification and using the
 post-transformation label as a function of attack strength, $\epsilon$.}
\label{fig:acc-vs-ep}
\end{figure}

\begin{figure}[b]
\centering
\includegraphics[width=\columnwidth]{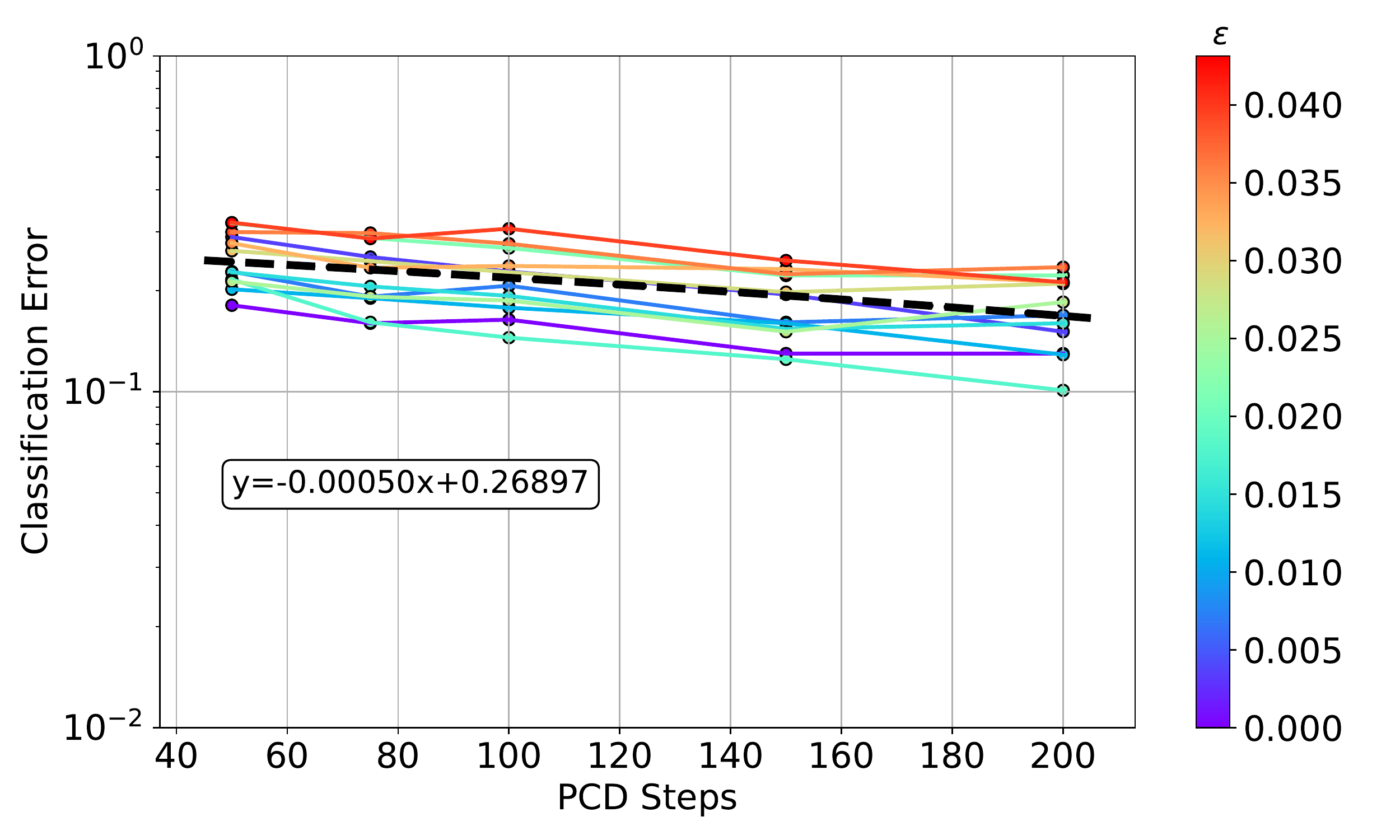}
\caption{Post-transformation PGD classification error as a function of the
 number of SGLD steps of each iteration of training. The color shade of the
 curves indicate the strength, $\epsilon$, of the applied PGD attack.}
\label{fig:error-vs-n}
\end{figure}

\begin{figure}[t]
\centering
\includegraphics[width=\columnwidth]{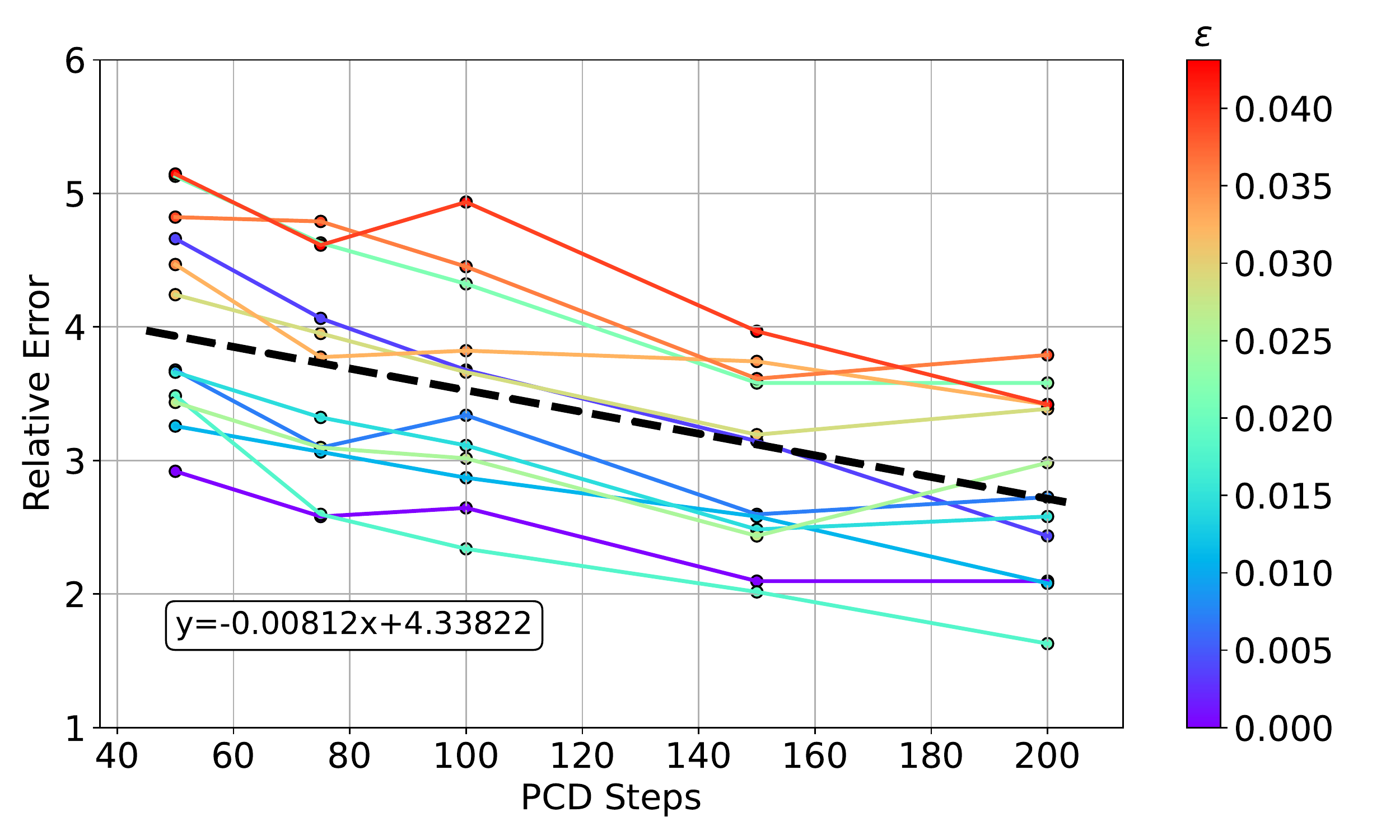}
\caption{Classification error relative to the performance of the baseline
 classifier on clean images. The color shade of the curves indicate the
 strength, $\epsilon$, of the applied PGD attack.}
\label{fig:rel-error-vs-n}
\end{figure}

An adversarial attack on a classifier with output logits $f(x) \in
\mathbb{R}^{|\mathcal Y|}$ searches for a sample $x \in \mathbb [0, 1]^d$ in a
neighborhood $S$ of an attacked image $x^+$ that maximizes the cross-entropy
loss:
\begin{equation}
x_{adv}(x^+,y) = \argmax_{x \in S} L(f(x),y).
\end{equation}
The cross-entropy loss $L(f(x), y)$ is $-\log(p_y)$ where $p_y$ is the
likelihood of the label $y$, obtained by passing $f(x)$ through the softmax
operator. $S$ is an $\epsilon$-ball around $x^+$ in the $L_\infty$-norm
and we have assumed pixel intensities between $0$ to $1$ for images.
The PGD attack scheme is summarized in \cref{fig:PGD-attack}.
The attack starts from a random initial point in $S$ and maximizes the above
objective function via iterative advances along a steepest descent direction
followed by projections back onto $S$. That is,
\begin{equation}
\begin{split}
x_{i+1} &= \Pi_{S}(x_i+\alpha g(x_i,y)), \quad \text{where}\\
g(x,y) &= \argmax_{||v||_\infty \leq 1} v^{\intercal}\Delta(x,y)
=\text{sign} (\Delta(x,y)).
\end{split}
\end{equation}
Here $\Pi_S$ denotes the projection onto $S$, $\alpha$ is the attack step size,
and $\Delta(x,y)$ is the attack gradient which is $\nabla_x L(f(x),y)$ in the
notation above.

We generate 1,000 adversarial images via PGD attacks on the
classifier for various attack strengths $\epsilon > 0$. These images are then
diffused through the EBMs using 1,500 Langevin steps for 150 parallel trials.
The outputs are provided to the classifier and the logits of all the trials are
averaged
\begin{equation}
F(x)\simeq \mathbb E_{T(x)}[f(T(x))]
\end{equation}
to provide a final \emph{post-transformation} label. Here $T(x)$ represents the
random tensor produced after diffusion of $x$ through Langevin dynamics. This
procedure, known as the expectation over transformation (EOT) defense
\citep{hill2020stochastic}, is depicted in \cref{fig:EOT}.

We perform this procedure using EBMs trained with
varying numbers of Langevin steps and demonstrate that longer Langevin
sampling improves the adversarial robustness of this defense.
As shown in \cref{fig:acc-vs-ep}, the standalone classifier is severely
vulnerable to PGD attacks, while energy-based purification restores most
adversarial images to their original classification labels. In
\cref{fig:error-vs-n,fig:rel-error-vs-n} we observe that an increase in the
duration of Langevin dynamics during EBM training confers greater
restorative power, augmenting the model's ability to distill adversarial
signals. We also regress an average exponential decay rate in the
classification error across various choices of $\epsilon$ and as a function
of the number of PCD steps.

\begin{table}[b]
    \centering
    \begin{tabular}{c|cccccccccccc}
    \hline
    $255 \times \epsilon$ & 0 & 1 & 2 & 3 & 4 & 5 & 6 & 7 & 8 & 9 & 10 & 11 \\
    $n$ & 338 & 300 & 450 & 348 & 380 & 247 & 422 & 655 & 553 & 651 & 503 & 419\\
    \hline
    \end{tabular}
    \caption{The projected number of SGLD step required for full purification.
     For each attack strength $\epsilon$, the reported value of $n$ indicates
     the predicted number of SGLD steps required during training of the model
     to restore the classifier back to its original performance with respect to
     classification of the PGD adversarial images according to regression of a
     linear trend between the relative classification error and the number of
     SGLD steps in \cref{fig:rel-error-vs-n}.}
    \label{tab:epsilon}
\end{table}

The slow rate of this decay suggests that achieving adversarial security may
require training EBMs with very large numbers of SGLD steps. For further
investigation, we also calculate the relative accuracy error of the defense
compared to the original classification error of the model on clean images.
Since the goal of purification is to recover the classifier's original accuracy
on clean images, the relative error is ideally 1 indicating the classifier
performs equally well on both clean and adversarial images post-transformation.
\cref{tab:epsilon} shows
the predicted number of PCD steps, according to the linear fit in
\cref{fig:rel-error-vs-n}, needed during training to allow the EBM to
restore the classifier's performance on post-transformation adversarial images
to that of the original classifier for each $\epsilon$. We note that a
projected runtime for a model utilizing $450$ Langevin steps during training in
the same fashion as our models is approximately $3.25$ days on a single
Nvidia Tesla V100S GPU for CIFAR-10 which is not a large dataset.

\begin{figure}[b]
\centering
\includegraphics[width=\columnwidth]{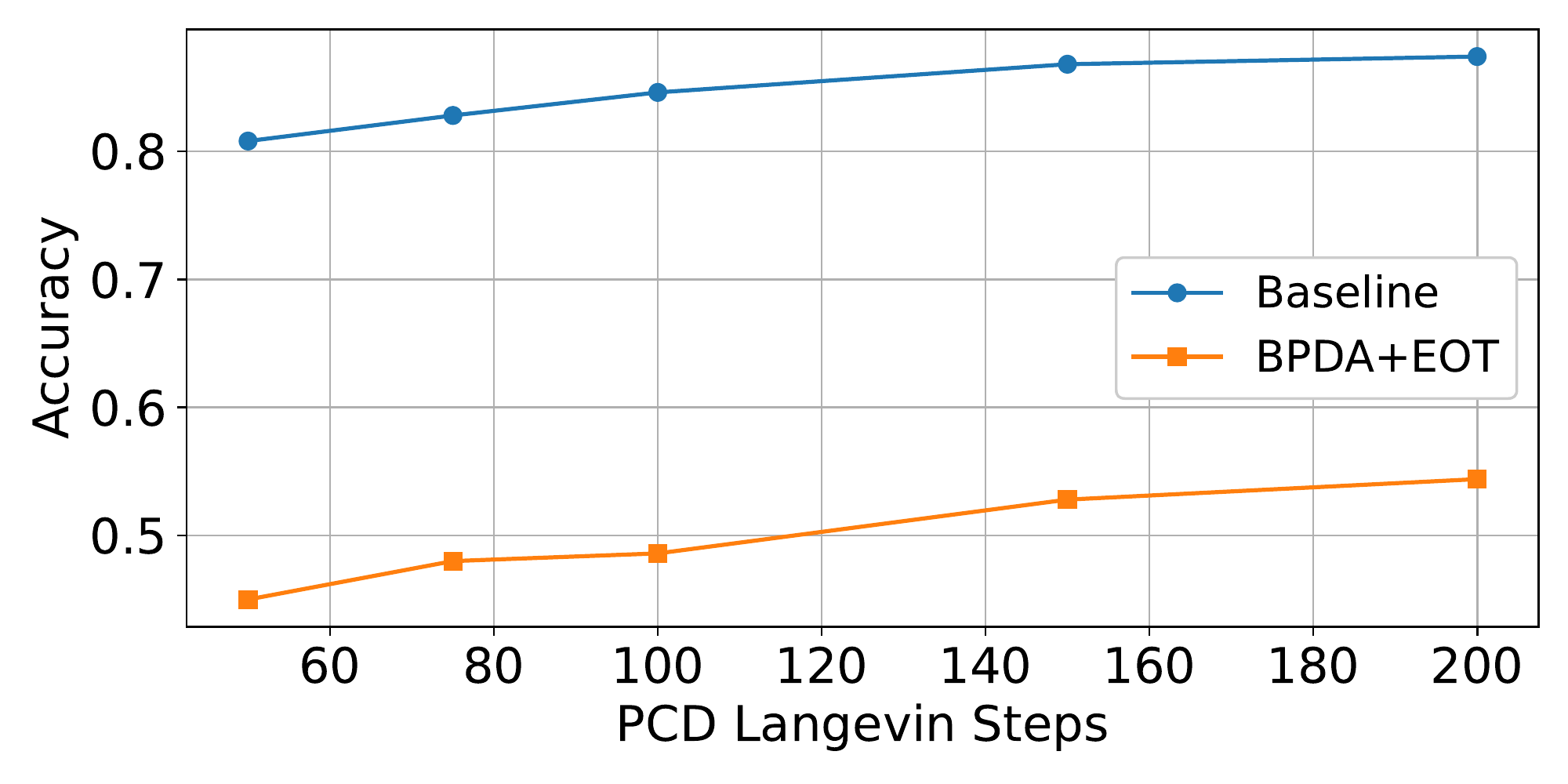}
\caption{Prediction accuracy of the EBM and the WRN classifier as a function
 of SGLD steps used during training after BPDA+EOT attacks. `Baseline' is the
 pre-attack accuracy, while the other curve shows the post-attack accuracy. The
 BPDA+EOT attack degrades the performance of the model, however, longer SGLD
 runs confer greater resistance to the attacks.}
\label{fig:BPDA_acc-vs-n}
\end{figure}

\begin{figure}[t]
\centering
\includegraphics[width=\columnwidth]{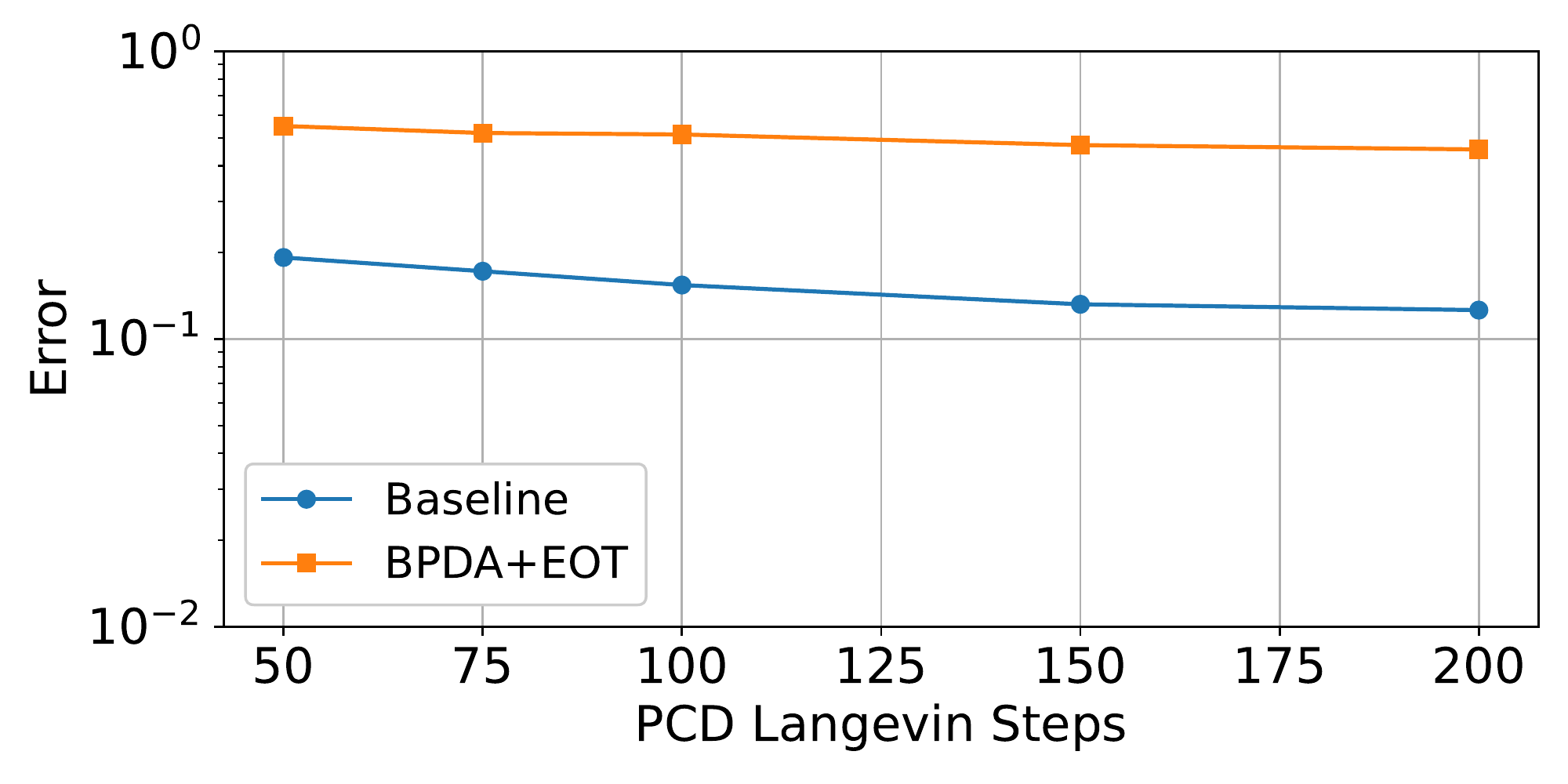}
\caption{Classification error after BPDA+EOT attacks in a logarithmic scale
 as a function of the number of SGLD steps used during training. `Baseline' is
 the pre-attack accuracy, while the other curve shows the post-attack
 accuracy.}
\label{fig:BPDA-error-vs-n}
\end{figure}

The EOT defense, itself, gives rise to one of the strongest attacks in the
literature known as the EOT attack \citep{athalye2018obfuscated}, which exploits
access to the classifier logits for a finite number $m$ of i.i.d. attack samples
$x_1, \ldots, x_m \sim T(x)$ and the following approximation of $F(x)$:
\begin{equation}
\hat{F}_{m}(x)= \frac{1}m \sum_{i=1}^m f(\hat x_i).
\end{equation}
Performing PGD attacks on these logits requires access to derivatives of the
diffusion process $T$ which is overcome using \emph{backward pass
differentiable approximation} (BPDA) \citep{athalye2018obfuscated} which in this
simplest case approximates the transformation $T$ via the identity mapping.
Therefore PGD will use $L(f(T(x), y)$ in the forward pass, but use $\nabla_x L(f
(x), y)$ as the attack gradient in the backward pass. Together, the EOT and BDPA
give rise to the following attack gradient for PGD:
\begin{equation}
\Delta (x,y) = \frac{1}{m} \sum_{i=1}^{m}
\nabla_{\hat{x}_i} L \bigg{(}\frac{1}{m} \sum_{i=1}^{m} f(\hat{x}_i),y\bigg{)}.
\end{equation}

Performing BPDA+EOT attacks is computationally expensive, and therefore the
results in \cref{fig:BPDA_acc-vs-n,fig:BPDA-error-vs-n}
are restricted to the fixed attack strength of
$\epsilon=8/255$ (a customary attack strength used in the literature as a
benchmark). For these attacks, we report a baseline accuracy and prediction
error that reflect the pre-attack performance of each EBM--classifier
combination. We observe that the baseline model's classification accuracy
improves with longer runs of Langevin dynamics during training. The results of
the BPDA+EOT attack are shown on the same plots. Although this attack
substantially reduces the classification performance of the joint model, EBMs
trained with longer-run Langevin dynamics still demonstrate greater
resistance.

\section{Calibration of the model}
\label{sec:calibration}

\begin{figure}[b]
\centering
\includegraphics[width=\columnwidth]{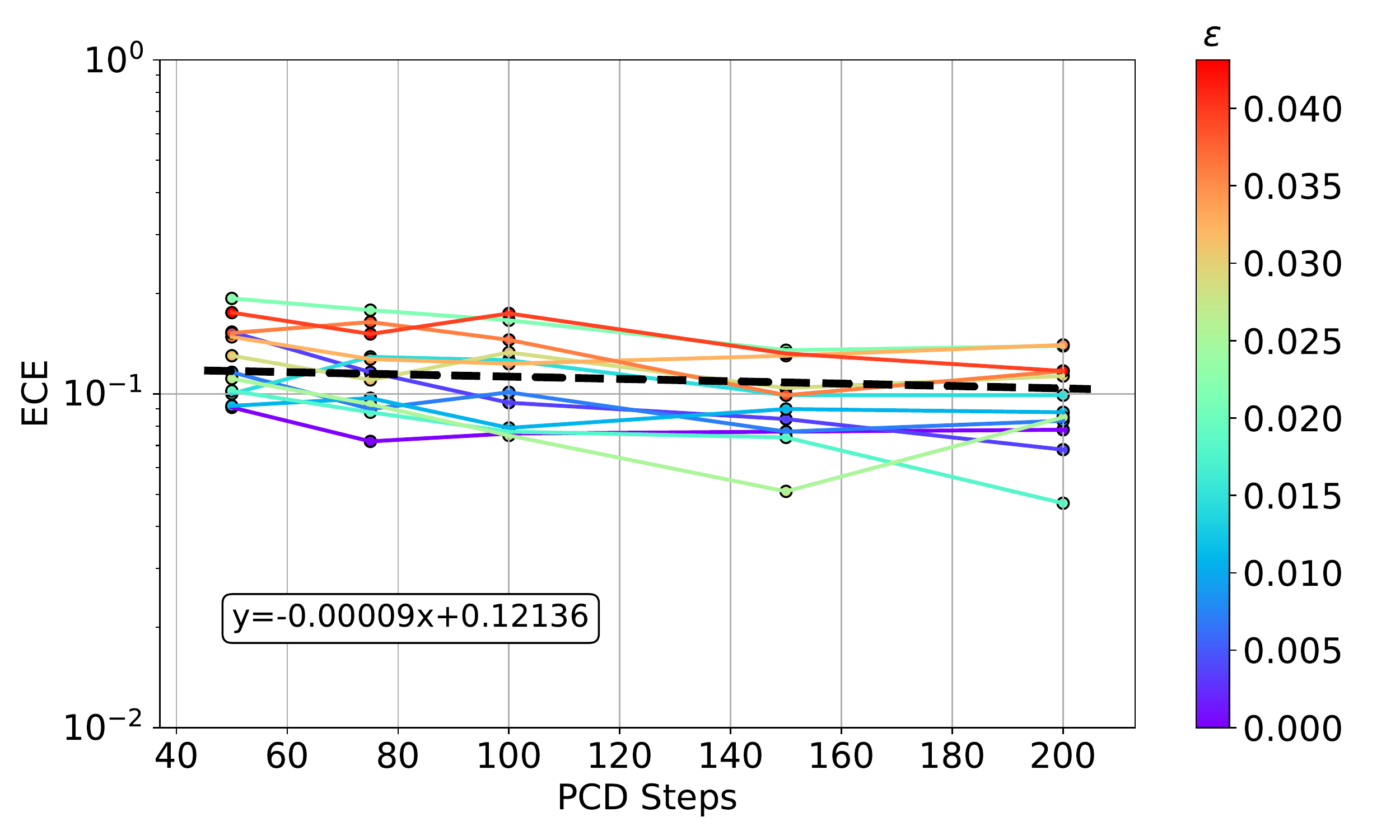}
\caption{Expected calibration error (ECE) as a function of the number of SGLD
steps executed during the training of EBMs. The ECE calculation pertains to
the post-transformation adversarial images. The color shade of the curves
shows the strength of the applied PGD attack.}
\label{fig:calib}
\end{figure}

In addition to improved adversarial robustness, we provide numerical
evidence that prolonged PCD during EBM training confer
improvements to the calibration of the model as well. Specifically, we study
the calibration in the classification of adversarial images, since calibration
also pertains to classification capabilities and the standalone EBM is not
itself a discriminative model. Calibration indicates that the model's
predictive confidence, $\max_y p(y|x)$, is commensurate with its
misclassification rate. To calculate the expected calibration error (ECE), we
find the classification confidence of each sample $x$ in the dataset, and
group them into equally spaced bins, $\{B_m\}^M_{m=1}$. For example, with
$M = 20$ bins, $B_0$ corresponds to all images for which the classifier's
confidence is between $0$ and $0.05$. The ECE is therefore defined as
\begin{equation}
\text{ECE} = \sum_{m=1}^M \frac{|B_m|}{N}|\text{acc}(B_m)-\text{conf}(B_m)|
\end{equation}
where $N$ is in the number of examples in the dataset, and $\text{acc}(B_m)$ and
$\text{conf}(B_m)$ are respectively the average accuracy and confidence for
the samples in bin $B_m$.

\cref{fig:calib} shows the ECE of post-purification adversarial images
against the number of PCD steps, for each PGD attack strength $\epsilon$.
The slope of the trend line is indicative of a gradual but steady improvement
in the calibration error as the number of PCD steps increases.

\section{Conclusion}

A key missing element on the path to fault-tolerant quantum computation is
practical utility measures that quantify how transformative the technology will
be \citep{darpa2021benchmark}. Quantum algorithms for solving differential
equations can be used to simulate diffusion processes of interest in generative
modeling
\citep{motamedi2022gibbs, du2019implicit, grathwohl2019your, ho2020denoising}.
Also, it has been shown that the models trained in this fashion exhibit
improved stochastic security \citep{du2019implicit, grathwohl2019your,
hill2020stochastic}. Therefore, in this paper we propose stochastic security as
a practical metric for gauging the utility of quantum Gibbs samplers in machine
learning tasks.

Because diffusion processes provide the mathematical models of equilibrium and
non-equilibrium thermodynamics, they play critical roles in machine learning.
The high computational cost of simulating these processes
classically---including time, energy, carbon, and water footprint of
the computation \citep{desislavov2021compute, li2023making}---has
motivated machine learning scientists to critically investigate the practical
benefits of models that rely on them.
In this paper we focused on deep energy-based models (EBM) as an example from
this family \citep{du2019implicit}. We examined two
specific utility measures, namely, the adversarial robustness and calibration
of EBMs. The EBMs were trained on classical image data using various
amounts of computational budget allocated for performing stochastic gradient
Langevin dynamics (SGLD) in order to simulate the diffusion process.
After training, diffusion of a test image using the energy potential represented
by the trained EBM is used as a transformation that purifies the image from the
adversarial attack before presenting it to a wide residual network (WRN)
classifier.

Our results demonstrate that purifying samples via EBMs trained with larger
numbers of SGLD steps improves the adversarial robustness of the WRN
classifier. Interestingly, the same diffusive transformation also yields
more calibrated classifications. We visually observe exponential decays in
adversarial vulnerability and calibration error of the WRN, and quantify these
trends using linear regression. Overall, our findings offer promising evidence
for the practical utility of continuous-domain quantum Gibbs samplers on
classical-data problems, all without relying on costly QRAMs.

\acknowledgments

This material is based upon work supported by the U.S.\ Department of Homeland
Security through a contract with the Critical Infrastructure Resilience
Institute (CIRI) at the University of Illinois. N.~A.~C., L.~S., and G.~S.\
acknowledge support by the U.S.\ Army Research Office under grant
W911NF-19-1-0397 and the U.S.\ National Science Foundation under grant
DGE-2152168. P.~R. further acknowledges the support of NSERC Discovery
grant RGPIN-2022-03339, Mike and Ophelia Lazaridis, Innovation, Science and
Economic Development Canada (ISED), 1QBit, and the Perimeter Institute for
Theoretical Physics. Research at the Perimeter Institute is supported in part
by the Government of Canada through ISED, and by the Province of Ontario
through the Ministry of Colleges and Universities. This research used resources
of the Oak Ridge Leadership Computing Facility, which is a DOE Office of Science
User Facility supported under Contract DE-AC05-00OR22725.

The authors thank Simon~Verret and Jeff~Hnybida for useful discussions.
The authors have no conflicts of interest to declare. All authors contributed to
the study conception and design. Numerical experiments, analysis of the
results, and the generation of figures were performed by N.~A.~C., and L.~S.
All authors read and approved the final manuscript.


\bibliography{main}

\begin{thebibliography}{50}%
\makeatletter
\providecommand \@ifxundefined [1]{%
 \@ifx{#1\undefined}
}%
\providecommand \@ifnum [1]{%
 \ifnum #1\expandafter \@firstoftwo
 \else \expandafter \@secondoftwo
 \fi
}%
\providecommand \@ifx [1]{%
 \ifx #1\expandafter \@firstoftwo
 \else \expandafter \@secondoftwo
 \fi
}%
\providecommand \natexlab [1]{#1}%
\providecommand \enquote  [1]{``#1''}%
\providecommand \bibnamefont  [1]{#1}%
\providecommand \bibfnamefont [1]{#1}%
\providecommand \citenamefont [1]{#1}%
\providecommand \href@noop [0]{\@secondoftwo}%
\providecommand \href [0]{\begingroup \@sanitize@url \@href}%
\providecommand \@href[1]{\@@startlink{#1}\@@href}%
\providecommand \@@href[1]{\endgroup#1\@@endlink}%
\providecommand \@sanitize@url [0]{\catcode `\\12\catcode `\$12\catcode
  `\&12\catcode `\#12\catcode `\^12\catcode `\_12\catcode `\%12\relax}%
\providecommand \@@startlink[1]{}%
\providecommand \@@endlink[0]{}%
\providecommand \url  [0]{\begingroup\@sanitize@url \@url }%
\providecommand \@url [1]{\endgroup\@href {#1}{\urlprefix }}%
\providecommand \urlprefix  [0]{URL }%
\providecommand \Eprint [0]{\href }%
\providecommand \doibase [0]{https://doi.org/}%
\providecommand \selectlanguage [0]{\@gobble}%
\providecommand \bibinfo  [0]{\@secondoftwo}%
\providecommand \bibfield  [0]{\@secondoftwo}%
\providecommand \translation [1]{[#1]}%
\providecommand \BibitemOpen [0]{}%
\providecommand \bibitemStop [0]{}%
\providecommand \bibitemNoStop [0]{.\EOS\space}%
\providecommand \EOS [0]{\spacefactor3000\relax}%
\providecommand \BibitemShut  [1]{\csname bibitem#1\endcsname}%
\let\auto@bib@innerbib\@empty
\bibitem [{\citenamefont {Ramesh}\ \emph {et~al.}(2022)\citenamefont {Ramesh},
  \citenamefont {Dhariwal}, \citenamefont {Nichol}, \citenamefont {Chu},\ and\
  \citenamefont {Chen}}]{ramesh2022hierarchical}%
  \BibitemOpen
  \bibfield  {author} {\bibinfo {author} {\bibfnamefont {A.}~\bibnamefont
  {Ramesh}}, \bibinfo {author} {\bibfnamefont {P.}~\bibnamefont {Dhariwal}},
  \bibinfo {author} {\bibfnamefont {A.}~\bibnamefont {Nichol}}, \bibinfo
  {author} {\bibfnamefont {C.}~\bibnamefont {Chu}},\ and\ \bibinfo {author}
  {\bibfnamefont {M.}~\bibnamefont {Chen}},\ }\bibfield  {title} {\bibinfo
  {title} {Hierarchical text-conditional image generation with clip latents},\
  }\href@noop {} {\bibfield  {journal} {\bibinfo  {journal} {arXiv preprint
  arXiv:2204.06125}\ } (\bibinfo {year} {2022})}\BibitemShut {NoStop}%
\bibitem [{\citenamefont {Rombach}\ \emph {et~al.}(2022)\citenamefont
  {Rombach}, \citenamefont {Blattmann}, \citenamefont {Lorenz}, \citenamefont
  {Esser},\ and\ \citenamefont {Ommer}}]{rombach2022high}%
  \BibitemOpen
  \bibfield  {author} {\bibinfo {author} {\bibfnamefont {R.}~\bibnamefont
  {Rombach}}, \bibinfo {author} {\bibfnamefont {A.}~\bibnamefont {Blattmann}},
  \bibinfo {author} {\bibfnamefont {D.}~\bibnamefont {Lorenz}}, \bibinfo
  {author} {\bibfnamefont {P.}~\bibnamefont {Esser}},\ and\ \bibinfo {author}
  {\bibfnamefont {B.}~\bibnamefont {Ommer}},\ }\bibfield  {title} {\bibinfo
  {title} {High-resolution image synthesis with latent diffusion models},\ }in\
  \href@noop {} {\emph {\bibinfo {booktitle} {Proceedings of the IEEE/CVF
  conference on computer vision and pattern recognition}}}\ (\bibinfo {year}
  {2022})\ pp.\ \bibinfo {pages} {10684--10695}\BibitemShut {NoStop}%
\bibitem [{\citenamefont {Brown}\ \emph {et~al.}(2020)\citenamefont {Brown},
  \citenamefont {Mann}, \citenamefont {Ryder}, \citenamefont {Subbiah},
  \citenamefont {Kaplan}, \citenamefont {Dhariwal}, \citenamefont
  {Neelakantan}, \citenamefont {Shyam}, \citenamefont {Sastry}, \citenamefont
  {Askell} \emph {et~al.}}]{brown2020language}%
  \BibitemOpen
  \bibfield  {author} {\bibinfo {author} {\bibfnamefont {T.}~\bibnamefont
  {Brown}}, \bibinfo {author} {\bibfnamefont {B.}~\bibnamefont {Mann}},
  \bibinfo {author} {\bibfnamefont {N.}~\bibnamefont {Ryder}}, \bibinfo
  {author} {\bibfnamefont {M.}~\bibnamefont {Subbiah}}, \bibinfo {author}
  {\bibfnamefont {J.~D.}\ \bibnamefont {Kaplan}}, \bibinfo {author}
  {\bibfnamefont {P.}~\bibnamefont {Dhariwal}}, \bibinfo {author}
  {\bibfnamefont {A.}~\bibnamefont {Neelakantan}}, \bibinfo {author}
  {\bibfnamefont {P.}~\bibnamefont {Shyam}}, \bibinfo {author} {\bibfnamefont
  {G.}~\bibnamefont {Sastry}}, \bibinfo {author} {\bibfnamefont
  {A.}~\bibnamefont {Askell}}, \emph {et~al.},\ }\bibfield  {title} {\bibinfo
  {title} {Language models are few-shot learners},\ }\href@noop {} {\bibfield
  {journal} {\bibinfo  {journal} {Advances in neural information processing
  systems}\ }\textbf {\bibinfo {volume} {33}},\ \bibinfo {pages} {1877}
  (\bibinfo {year} {2020})}\BibitemShut {NoStop}%
\bibitem [{\citenamefont {OpenAI}(2023)}]{openai2023gpt4}%
  \BibitemOpen
  \bibfield  {author} {\bibinfo {author} {\bibnamefont {OpenAI}},\ }\href@noop
  {} {\bibinfo {title} {{GPT}-4 technical report}} (\bibinfo {year} {2023}),\
  \Eprint {https://arxiv.org/abs/2303.08774} {arXiv:2303.08774 [cs.CL]}
  \BibitemShut {NoStop}%
\bibitem [{\citenamefont {Motamedi}\ and\ \citenamefont
  {Ronagh}(2024)}]{motamedi2022gibbs}%
  \BibitemOpen
  \bibfield  {author} {\bibinfo {author} {\bibfnamefont {A.}~\bibnamefont
  {Motamedi}}\ and\ \bibinfo {author} {\bibfnamefont {P.}~\bibnamefont
  {Ronagh}},\ }\bibfield  {title} {\bibinfo {title} {Gibbs sampling of
  continuous potentials on a quantum computer},\ }\href
  {https://proceedings.mlr.press/v235/motamedi24a.html} {\bibfield  {journal}
  {\bibinfo  {journal} {Proceedings of Machine Learning Research}\ }\textbf
  {\bibinfo {volume} {235}},\ \bibinfo {pages} {36322} (\bibinfo {year}
  {2024})}\BibitemShut {NoStop}%
\bibitem [{\citenamefont {Du}\ and\ \citenamefont
  {Mordatch}(2019)}]{du2019implicit}%
  \BibitemOpen
  \bibfield  {author} {\bibinfo {author} {\bibfnamefont {Y.}~\bibnamefont
  {Du}}\ and\ \bibinfo {author} {\bibfnamefont {I.}~\bibnamefont {Mordatch}},\
  }\bibfield  {title} {\bibinfo {title} {Implicit generation and generalization
  in energy-based models},\ }\href@noop {} {\bibfield  {journal} {\bibinfo
  {journal} {arXiv preprint arXiv:1903.08689}\ } (\bibinfo {year}
  {2019})}\BibitemShut {NoStop}%
\bibitem [{\citenamefont {Grathwohl}\ \emph {et~al.}(2019)\citenamefont
  {Grathwohl}, \citenamefont {Wang}, \citenamefont {Jacobsen}, \citenamefont
  {Duvenaud}, \citenamefont {Norouzi},\ and\ \citenamefont
  {Swersky}}]{grathwohl2019your}%
  \BibitemOpen
  \bibfield  {author} {\bibinfo {author} {\bibfnamefont {W.}~\bibnamefont
  {Grathwohl}}, \bibinfo {author} {\bibfnamefont {K.-C.}\ \bibnamefont {Wang}},
  \bibinfo {author} {\bibfnamefont {J.-H.}\ \bibnamefont {Jacobsen}}, \bibinfo
  {author} {\bibfnamefont {D.}~\bibnamefont {Duvenaud}}, \bibinfo {author}
  {\bibfnamefont {M.}~\bibnamefont {Norouzi}},\ and\ \bibinfo {author}
  {\bibfnamefont {K.}~\bibnamefont {Swersky}},\ }\bibfield  {title} {\bibinfo
  {title} {Your classifier is secretly an energy based model and you should
  treat it like one},\ }\href@noop {} {\bibfield  {journal} {\bibinfo
  {journal} {arXiv preprint arXiv:1912.03263}\ } (\bibinfo {year}
  {2019})}\BibitemShut {NoStop}%
\bibitem [{\citenamefont {Song}\ and\ \citenamefont
  {Kingma}(2021)}]{song2021train}%
  \BibitemOpen
  \bibfield  {author} {\bibinfo {author} {\bibfnamefont {Y.}~\bibnamefont
  {Song}}\ and\ \bibinfo {author} {\bibfnamefont {D.~P.}\ \bibnamefont
  {Kingma}},\ }\bibfield  {title} {\bibinfo {title} {How to train your
  energy-based models},\ }\href@noop {} {\bibfield  {journal} {\bibinfo
  {journal} {arXiv preprint arXiv:2101.03288}\ } (\bibinfo {year}
  {2021})}\BibitemShut {NoStop}%
\bibitem [{\citenamefont {Hill}\ \emph {et~al.}(2020)\citenamefont {Hill},
  \citenamefont {Mitchell},\ and\ \citenamefont {Zhu}}]{hill2020stochastic}%
  \BibitemOpen
  \bibfield  {author} {\bibinfo {author} {\bibfnamefont {M.}~\bibnamefont
  {Hill}}, \bibinfo {author} {\bibfnamefont {J.}~\bibnamefont {Mitchell}},\
  and\ \bibinfo {author} {\bibfnamefont {S.-C.}\ \bibnamefont {Zhu}},\
  }\bibfield  {title} {\bibinfo {title} {Stochastic security: Adversarial
  defense using long-run dynamics of energy-based models},\ }\href@noop {}
  {\bibfield  {journal} {\bibinfo  {journal} {arXiv preprint arXiv:2005.13525}\
  } (\bibinfo {year} {2020})}\BibitemShut {NoStop}%
\bibitem [{\citenamefont {Guo}\ \emph {et~al.}(2017)\citenamefont {Guo},
  \citenamefont {Pleiss}, \citenamefont {Sun},\ and\ \citenamefont
  {Weinberger}}]{guo2017calibration}%
  \BibitemOpen
  \bibfield  {author} {\bibinfo {author} {\bibfnamefont {C.}~\bibnamefont
  {Guo}}, \bibinfo {author} {\bibfnamefont {G.}~\bibnamefont {Pleiss}},
  \bibinfo {author} {\bibfnamefont {Y.}~\bibnamefont {Sun}},\ and\ \bibinfo
  {author} {\bibfnamefont {K.~Q.}\ \bibnamefont {Weinberger}},\ }\bibfield
  {title} {\bibinfo {title} {On calibration of modern neural networks},\ }in\
  \href@noop {} {\emph {\bibinfo {booktitle} {International Conference on
  Machine Learning}}}\ (\bibinfo {organization} {PMLR},\ \bibinfo {year}
  {2017})\ pp.\ \bibinfo {pages} {1321--1330}\BibitemShut {NoStop}%
\bibitem [{\citenamefont {Zagoruyko}\ and\ \citenamefont
  {Komodakis}(2016)}]{Zagoruyko2016}%
  \BibitemOpen
  \bibfield  {author} {\bibinfo {author} {\bibfnamefont {S.}~\bibnamefont
  {Zagoruyko}}\ and\ \bibinfo {author} {\bibfnamefont {N.}~\bibnamefont
  {Komodakis}},\ }\href@noop {} {\bibinfo {title} {Wide residual networks}}
  (\bibinfo {year} {2016})\BibitemShut {NoStop}%
\bibitem [{\citenamefont {Ackley}\ \emph {et~al.}(1985)\citenamefont {Ackley},
  \citenamefont {Hinton},\ and\ \citenamefont
  {Sejnowski}}]{ackley1985learning}%
  \BibitemOpen
  \bibfield  {author} {\bibinfo {author} {\bibfnamefont {D.~H.}\ \bibnamefont
  {Ackley}}, \bibinfo {author} {\bibfnamefont {G.~E.}\ \bibnamefont {Hinton}},\
  and\ \bibinfo {author} {\bibfnamefont {T.~J.}\ \bibnamefont {Sejnowski}},\
  }\bibfield  {title} {\bibinfo {title} {A learning algorithm for {B}oltzmann
  machines},\ }\href@noop {} {\bibfield  {journal} {\bibinfo  {journal}
  {Cognitive science}\ }\textbf {\bibinfo {volume} {9}},\ \bibinfo {pages}
  {147} (\bibinfo {year} {1985})}\BibitemShut {NoStop}%
\bibitem [{\citenamefont {Hopfield}(1982)}]{hopfield1982neural}%
  \BibitemOpen
  \bibfield  {author} {\bibinfo {author} {\bibfnamefont {J.~J.}\ \bibnamefont
  {Hopfield}},\ }\bibfield  {title} {\bibinfo {title} {Neural networks and
  physical systems with emergent collective computational abilities},\
  }\href@noop {} {\bibfield  {journal} {\bibinfo  {journal} {Proceedings of the
  national academy of sciences}\ }\textbf {\bibinfo {volume} {79}},\ \bibinfo
  {pages} {2554} (\bibinfo {year} {1982})}\BibitemShut {NoStop}%
\bibitem [{\citenamefont {Hinton}\ \emph {et~al.}(2006)\citenamefont {Hinton},
  \citenamefont {Osindero},\ and\ \citenamefont {Teh}}]{hinton2006fast}%
  \BibitemOpen
  \bibfield  {author} {\bibinfo {author} {\bibfnamefont {G.~E.}\ \bibnamefont
  {Hinton}}, \bibinfo {author} {\bibfnamefont {S.}~\bibnamefont {Osindero}},\
  and\ \bibinfo {author} {\bibfnamefont {Y.-W.}\ \bibnamefont {Teh}},\
  }\bibfield  {title} {\bibinfo {title} {A fast learning algorithm for deep
  belief nets},\ }\href@noop {} {\bibfield  {journal} {\bibinfo  {journal}
  {Neural computation}\ }\textbf {\bibinfo {volume} {18}},\ \bibinfo {pages}
  {1527} (\bibinfo {year} {2006})}\BibitemShut {NoStop}%
\bibitem [{\citenamefont {Hinton}(2007)}]{hinton2007learning}%
  \BibitemOpen
  \bibfield  {author} {\bibinfo {author} {\bibfnamefont {G.~E.}\ \bibnamefont
  {Hinton}},\ }\bibfield  {title} {\bibinfo {title} {Learning multiple layers
  of representation},\ }\href@noop {} {\bibfield  {journal} {\bibinfo
  {journal} {Trends in cognitive sciences}\ }\textbf {\bibinfo {volume} {11}},\
  \bibinfo {pages} {428} (\bibinfo {year} {2007})}\BibitemShut {NoStop}%
\bibitem [{\citenamefont {Koller}\ and\ \citenamefont
  {Friedman}(2009)}]{koller2009probabilistic}%
  \BibitemOpen
  \bibfield  {author} {\bibinfo {author} {\bibfnamefont {D.}~\bibnamefont
  {Koller}}\ and\ \bibinfo {author} {\bibfnamefont {N.}~\bibnamefont
  {Friedman}},\ }\href@noop {} {\emph {\bibinfo {title} {Probabilistic
  graphical models: principles and techniques}}}\ (\bibinfo  {publisher} {MIT
  press},\ \bibinfo {address} {Cambridge, Massachusetts},\ \bibinfo {year}
  {2009})\BibitemShut {NoStop}%
\bibitem [{\citenamefont {Clifford}\ and\ \citenamefont
  {Hammersley}(1971)}]{clifford1971markov}%
  \BibitemOpen
  \bibfield  {author} {\bibinfo {author} {\bibfnamefont {P.}~\bibnamefont
  {Clifford}}\ and\ \bibinfo {author} {\bibfnamefont {J.}~\bibnamefont
  {Hammersley}},\ }\bibfield  {title} {\bibinfo {title} {Markov fields on
  finite graphs and lattices},\ }\href@noop {} {\  (\bibinfo {year}
  {1971})}\BibitemShut {NoStop}%
\bibitem [{\citenamefont {Nijkamp}\ \emph {et~al.}(2020)\citenamefont
  {Nijkamp}, \citenamefont {Hill}, \citenamefont {Han}, \citenamefont {Zhu},\
  and\ \citenamefont {Wu}}]{nijkamp2020anatomy}%
  \BibitemOpen
  \bibfield  {author} {\bibinfo {author} {\bibfnamefont {E.}~\bibnamefont
  {Nijkamp}}, \bibinfo {author} {\bibfnamefont {M.}~\bibnamefont {Hill}},
  \bibinfo {author} {\bibfnamefont {T.}~\bibnamefont {Han}}, \bibinfo {author}
  {\bibfnamefont {S.-C.}\ \bibnamefont {Zhu}},\ and\ \bibinfo {author}
  {\bibfnamefont {Y.~N.}\ \bibnamefont {Wu}},\ }\bibfield  {title} {\bibinfo
  {title} {On the anatomy of {MCMC}-based maximum likelihood learning of
  energy-based models},\ }in\ \href@noop {} {\emph {\bibinfo {booktitle}
  {Proceedings of the AAAI Conference on Artificial Intelligence}}},\
  Vol.~\bibinfo {volume} {34}\ (\bibinfo {year} {2020})\ pp.\ \bibinfo {pages}
  {5272--5280}\BibitemShut {NoStop}%
\bibitem [{\citenamefont {Che}\ \emph {et~al.}(2020)\citenamefont {Che},
  \citenamefont {Zhang}, \citenamefont {Sohl-Dickstein}, \citenamefont
  {Larochelle}, \citenamefont {Paull}, \citenamefont {Cao},\ and\ \citenamefont
  {Bengio}}]{che2020your}%
  \BibitemOpen
  \bibfield  {author} {\bibinfo {author} {\bibfnamefont {T.}~\bibnamefont
  {Che}}, \bibinfo {author} {\bibfnamefont {R.}~\bibnamefont {Zhang}}, \bibinfo
  {author} {\bibfnamefont {J.}~\bibnamefont {Sohl-Dickstein}}, \bibinfo
  {author} {\bibfnamefont {H.}~\bibnamefont {Larochelle}}, \bibinfo {author}
  {\bibfnamefont {L.}~\bibnamefont {Paull}}, \bibinfo {author} {\bibfnamefont
  {Y.}~\bibnamefont {Cao}},\ and\ \bibinfo {author} {\bibfnamefont
  {Y.}~\bibnamefont {Bengio}},\ }\bibfield  {title} {\bibinfo {title} {Your
  {GAN} is secretly an energy-based model and you should use discriminator
  driven latent sampling},\ }\href@noop {} {\bibfield  {journal} {\bibinfo
  {journal} {Advances in Neural Information Processing Systems}\ }\textbf
  {\bibinfo {volume} {33}},\ \bibinfo {pages} {12275} (\bibinfo {year}
  {2020})}\BibitemShut {NoStop}%
\bibitem [{\citenamefont {Chen}\ \emph {et~al.}(2014)\citenamefont {Chen},
  \citenamefont {Fox},\ and\ \citenamefont {Guestrin}}]{chen2014stochastic}%
  \BibitemOpen
  \bibfield  {author} {\bibinfo {author} {\bibfnamefont {T.}~\bibnamefont
  {Chen}}, \bibinfo {author} {\bibfnamefont {E.}~\bibnamefont {Fox}},\ and\
  \bibinfo {author} {\bibfnamefont {C.}~\bibnamefont {Guestrin}},\ }\bibfield
  {title} {\bibinfo {title} {Stochastic gradient {H}amiltonian {M}onte
  {C}arlo},\ }in\ \href@noop {} {\emph {\bibinfo {booktitle} {International
  conference on machine learning}}}\ (\bibinfo {organization} {PMLR},\ \bibinfo
  {year} {2014})\ pp.\ \bibinfo {pages} {1683--1691}\BibitemShut {NoStop}%
\bibitem [{\citenamefont {Nijkamp}\ \emph {et~al.}(2019)\citenamefont
  {Nijkamp}, \citenamefont {Hill}, \citenamefont {Zhu},\ and\ \citenamefont
  {Wu}}]{nijkamp2019learning}%
  \BibitemOpen
  \bibfield  {author} {\bibinfo {author} {\bibfnamefont {E.}~\bibnamefont
  {Nijkamp}}, \bibinfo {author} {\bibfnamefont {M.}~\bibnamefont {Hill}},
  \bibinfo {author} {\bibfnamefont {S.-C.}\ \bibnamefont {Zhu}},\ and\ \bibinfo
  {author} {\bibfnamefont {Y.~N.}\ \bibnamefont {Wu}},\ }\bibfield  {title}
  {\bibinfo {title} {Learning non-convergent non-persistent short-run {MCMC}
  toward energy-based model},\ }\href@noop {} {\bibfield  {journal} {\bibinfo
  {journal} {Advances in Neural Information Processing Systems}\ }\textbf
  {\bibinfo {volume} {32}} (\bibinfo {year} {2019})}\BibitemShut {NoStop}%
\bibitem [{\citenamefont {Terhal}\ and\ \citenamefont
  {DiVincenzo}(2000)}]{terhal2000problem}%
  \BibitemOpen
  \bibfield  {author} {\bibinfo {author} {\bibfnamefont {B.~M.}\ \bibnamefont
  {Terhal}}\ and\ \bibinfo {author} {\bibfnamefont {D.~P.}\ \bibnamefont
  {DiVincenzo}},\ }\bibfield  {title} {\bibinfo {title} {Problem of
  equilibration and the computation of correlation functions on a quantum
  computer},\ }\href@noop {} {\bibfield  {journal} {\bibinfo  {journal}
  {Physical Review A}\ }\textbf {\bibinfo {volume} {61}},\ \bibinfo {pages}
  {022301} (\bibinfo {year} {2000})}\BibitemShut {NoStop}%
\bibitem [{\citenamefont {Poulin}\ and\ \citenamefont
  {Wocjan}(2009)}]{poulin2009sampling}%
  \BibitemOpen
  \bibfield  {author} {\bibinfo {author} {\bibfnamefont {D.}~\bibnamefont
  {Poulin}}\ and\ \bibinfo {author} {\bibfnamefont {P.}~\bibnamefont
  {Wocjan}},\ }\bibfield  {title} {\bibinfo {title} {Sampling from the thermal
  quantum {G}ibbs state and evaluating partition functions with a quantum
  computer},\ }\href@noop {} {\bibfield  {journal} {\bibinfo  {journal}
  {Physical review letters}\ }\textbf {\bibinfo {volume} {103}},\ \bibinfo
  {pages} {220502} (\bibinfo {year} {2009})}\BibitemShut {NoStop}%
\bibitem [{\citenamefont {Temme}\ \emph {et~al.}(2011)\citenamefont {Temme},
  \citenamefont {Osborne}, \citenamefont {Vollbrecht}, \citenamefont {Poulin},\
  and\ \citenamefont {Verstraete}}]{temme2011quantum}%
  \BibitemOpen
  \bibfield  {author} {\bibinfo {author} {\bibfnamefont {K.}~\bibnamefont
  {Temme}}, \bibinfo {author} {\bibfnamefont {T.~J.}\ \bibnamefont {Osborne}},
  \bibinfo {author} {\bibfnamefont {K.~G.}\ \bibnamefont {Vollbrecht}},
  \bibinfo {author} {\bibfnamefont {D.}~\bibnamefont {Poulin}},\ and\ \bibinfo
  {author} {\bibfnamefont {F.}~\bibnamefont {Verstraete}},\ }\bibfield  {title}
  {\bibinfo {title} {Quantum {M}etropolis sampling},\ }\href@noop {} {\bibfield
   {journal} {\bibinfo  {journal} {Nature}\ }\textbf {\bibinfo {volume}
  {471}},\ \bibinfo {pages} {87} (\bibinfo {year} {2011})}\BibitemShut
  {NoStop}%
\bibitem [{\citenamefont {Kastoryano}\ and\ \citenamefont
  {Brandao}(2016)}]{kastoryano2016quantum}%
  \BibitemOpen
  \bibfield  {author} {\bibinfo {author} {\bibfnamefont {M.~J.}\ \bibnamefont
  {Kastoryano}}\ and\ \bibinfo {author} {\bibfnamefont {F.~G.}\ \bibnamefont
  {Brandao}},\ }\bibfield  {title} {\bibinfo {title} {Quantum {G}ibbs samplers:
  the commuting case},\ }\href@noop {} {\bibfield  {journal} {\bibinfo
  {journal} {Communications in Mathematical Physics}\ }\textbf {\bibinfo
  {volume} {344}},\ \bibinfo {pages} {915} (\bibinfo {year}
  {2016})}\BibitemShut {NoStop}%
\bibitem [{\citenamefont {Chowdhury}\ and\ \citenamefont
  {Somma}(2017)}]{chowdhury2016quantum}%
  \BibitemOpen
  \bibfield  {author} {\bibinfo {author} {\bibfnamefont {A.~N.}\ \bibnamefont
  {Chowdhury}}\ and\ \bibinfo {author} {\bibfnamefont {R.~D.}\ \bibnamefont
  {Somma}},\ }\bibfield  {title} {\bibinfo {title} {Quantum algorithms for
  {G}ibbs sampling and hitting-time estimation},\ }\href@noop {} {\bibfield
  {journal} {\bibinfo  {journal} {Quantum Info. Comput.}\ }\textbf {\bibinfo
  {volume} {17}},\ \bibinfo {pages} {41–64} (\bibinfo {year}
  {2017})}\BibitemShut {NoStop}%
\bibitem [{\citenamefont {Van~Apeldoorn}\ \emph {et~al.}(2020)\citenamefont
  {Van~Apeldoorn}, \citenamefont {Gily{\'e}n}, \citenamefont {Gribling},\ and\
  \citenamefont {de~Wolf}}]{van2020quantum}%
  \BibitemOpen
  \bibfield  {author} {\bibinfo {author} {\bibfnamefont {J.}~\bibnamefont
  {Van~Apeldoorn}}, \bibinfo {author} {\bibfnamefont {A.}~\bibnamefont
  {Gily{\'e}n}}, \bibinfo {author} {\bibfnamefont {S.}~\bibnamefont
  {Gribling}},\ and\ \bibinfo {author} {\bibfnamefont {R.}~\bibnamefont
  {de~Wolf}},\ }\bibfield  {title} {\bibinfo {title} {Quantum {SDP}-solvers:
  Better upper and lower bounds},\ }\href@noop {} {\bibfield  {journal}
  {\bibinfo  {journal} {Quantum}\ }\textbf {\bibinfo {volume} {4}},\ \bibinfo
  {pages} {230} (\bibinfo {year} {2020})}\BibitemShut {NoStop}%
\bibitem [{\citenamefont {Lemieux}\ \emph {et~al.}(2020)\citenamefont
  {Lemieux}, \citenamefont {Heim}, \citenamefont {Poulin}, \citenamefont
  {Svore},\ and\ \citenamefont {Troyer}}]{lemieux2020efficient}%
  \BibitemOpen
  \bibfield  {author} {\bibinfo {author} {\bibfnamefont {J.}~\bibnamefont
  {Lemieux}}, \bibinfo {author} {\bibfnamefont {B.}~\bibnamefont {Heim}},
  \bibinfo {author} {\bibfnamefont {D.}~\bibnamefont {Poulin}}, \bibinfo
  {author} {\bibfnamefont {K.}~\bibnamefont {Svore}},\ and\ \bibinfo {author}
  {\bibfnamefont {M.}~\bibnamefont {Troyer}},\ }\bibfield  {title} {\bibinfo
  {title} {Efficient quantum walk circuits for {M}etropolis-{H}astings
  algorithm},\ }\href@noop {} {\bibfield  {journal} {\bibinfo  {journal}
  {Quantum}\ }\textbf {\bibinfo {volume} {4}},\ \bibinfo {pages} {287}
  (\bibinfo {year} {2020})}\BibitemShut {NoStop}%
\bibitem [{\citenamefont {Bravyi}\ \emph {et~al.}(2021)\citenamefont {Bravyi},
  \citenamefont {Chowdhury}, \citenamefont {Gosset},\ and\ \citenamefont
  {Wocjan}}]{bravyi2021complexity}%
  \BibitemOpen
  \bibfield  {author} {\bibinfo {author} {\bibfnamefont {S.}~\bibnamefont
  {Bravyi}}, \bibinfo {author} {\bibfnamefont {A.}~\bibnamefont {Chowdhury}},
  \bibinfo {author} {\bibfnamefont {D.}~\bibnamefont {Gosset}},\ and\ \bibinfo
  {author} {\bibfnamefont {P.}~\bibnamefont {Wocjan}},\ }\bibfield  {title}
  {\bibinfo {title} {On the complexity of quantum partition functions},\
  }\href@noop {} {\bibfield  {journal} {\bibinfo  {journal} {arXiv preprint
  arXiv:2110.15466}\ } (\bibinfo {year} {2021})}\BibitemShut {NoStop}%
\bibitem [{\citenamefont {Wiebe}\ \emph {et~al.}(2014)\citenamefont {Wiebe},
  \citenamefont {Kapoor},\ and\ \citenamefont {Svore}}]{wiebe2014quantum}%
  \BibitemOpen
  \bibfield  {author} {\bibinfo {author} {\bibfnamefont {N.}~\bibnamefont
  {Wiebe}}, \bibinfo {author} {\bibfnamefont {A.}~\bibnamefont {Kapoor}},\ and\
  \bibinfo {author} {\bibfnamefont {K.~M.}\ \bibnamefont {Svore}},\ }\bibfield
  {title} {\bibinfo {title} {Quantum deep learning},\ }\href@noop {} {\bibfield
   {journal} {\bibinfo  {journal} {arXiv preprint arXiv:1412.3489}\ } (\bibinfo
  {year} {2014})}\BibitemShut {NoStop}%
\bibitem [{\citenamefont {Amin}\ \emph {et~al.}(2018)\citenamefont {Amin},
  \citenamefont {Andriyash}, \citenamefont {Rolfe}, \citenamefont
  {Kulchytskyy},\ and\ \citenamefont {Melko}}]{amin2018quantum}%
  \BibitemOpen
  \bibfield  {author} {\bibinfo {author} {\bibfnamefont {M.~H.}\ \bibnamefont
  {Amin}}, \bibinfo {author} {\bibfnamefont {E.}~\bibnamefont {Andriyash}},
  \bibinfo {author} {\bibfnamefont {J.}~\bibnamefont {Rolfe}}, \bibinfo
  {author} {\bibfnamefont {B.}~\bibnamefont {Kulchytskyy}},\ and\ \bibinfo
  {author} {\bibfnamefont {R.}~\bibnamefont {Melko}},\ }\bibfield  {title}
  {\bibinfo {title} {Quantum {B}oltzmann machine},\ }\href@noop {} {\bibfield
  {journal} {\bibinfo  {journal} {Physical Review X}\ }\textbf {\bibinfo
  {volume} {8}},\ \bibinfo {pages} {021050} (\bibinfo {year}
  {2018})}\BibitemShut {NoStop}%
\bibitem [{\citenamefont {Crawford}\ \emph {et~al.}(2016)\citenamefont
  {Crawford}, \citenamefont {Levit}, \citenamefont {Ghadermarzy}, \citenamefont
  {Oberoi},\ and\ \citenamefont {Ronagh}}]{crawford2016reinforcement}%
  \BibitemOpen
  \bibfield  {author} {\bibinfo {author} {\bibfnamefont {D.}~\bibnamefont
  {Crawford}}, \bibinfo {author} {\bibfnamefont {A.}~\bibnamefont {Levit}},
  \bibinfo {author} {\bibfnamefont {N.}~\bibnamefont {Ghadermarzy}}, \bibinfo
  {author} {\bibfnamefont {J.~S.}\ \bibnamefont {Oberoi}},\ and\ \bibinfo
  {author} {\bibfnamefont {P.}~\bibnamefont {Ronagh}},\ }\bibfield  {title}
  {\bibinfo {title} {Reinforcement learning using quantum {B}oltzmann
  machines},\ }\href@noop {} {\bibfield  {journal} {\bibinfo  {journal} {arXiv
  preprint arXiv:1612.05695}\ } (\bibinfo {year} {2016})}\BibitemShut {NoStop}%
\bibitem [{\citenamefont {Levit}\ \emph {et~al.}(2017)\citenamefont {Levit},
  \citenamefont {Crawford}, \citenamefont {Ghadermarzy}, \citenamefont
  {Oberoi}, \citenamefont {Zahedinejad},\ and\ \citenamefont
  {Ronagh}}]{levit2017free}%
  \BibitemOpen
  \bibfield  {author} {\bibinfo {author} {\bibfnamefont {A.}~\bibnamefont
  {Levit}}, \bibinfo {author} {\bibfnamefont {D.}~\bibnamefont {Crawford}},
  \bibinfo {author} {\bibfnamefont {N.}~\bibnamefont {Ghadermarzy}}, \bibinfo
  {author} {\bibfnamefont {J.~S.}\ \bibnamefont {Oberoi}}, \bibinfo {author}
  {\bibfnamefont {E.}~\bibnamefont {Zahedinejad}},\ and\ \bibinfo {author}
  {\bibfnamefont {P.}~\bibnamefont {Ronagh}},\ }\bibfield  {title} {\bibinfo
  {title} {Free energy-based reinforcement learning using a quantum
  processor},\ }\href@noop {} {\bibfield  {journal} {\bibinfo  {journal} {arXiv
  preprint arXiv:1706.00074}\ } (\bibinfo {year} {2017})}\BibitemShut {NoStop}%
\bibitem [{\citenamefont {Sepehry}\ \emph {et~al.}(2022)\citenamefont
  {Sepehry}, \citenamefont {Iranmanesh}, \citenamefont {Friedlander},\ and\
  \citenamefont {Ronagh}}]{sepehry2022quantum}%
  \BibitemOpen
  \bibfield  {author} {\bibinfo {author} {\bibfnamefont {B.}~\bibnamefont
  {Sepehry}}, \bibinfo {author} {\bibfnamefont {E.}~\bibnamefont {Iranmanesh}},
  \bibinfo {author} {\bibfnamefont {M.~P.}\ \bibnamefont {Friedlander}},\ and\
  \bibinfo {author} {\bibfnamefont {P.}~\bibnamefont {Ronagh}},\ }\bibfield
  {title} {\bibinfo {title} {Quantum algorithms for structured prediction},\
  }\href@noop {} {\bibfield  {journal} {\bibinfo  {journal} {Quantum Machine
  Intelligence}\ }\textbf {\bibinfo {volume} {4}},\ \bibinfo {pages} {25}
  (\bibinfo {year} {2022})}\BibitemShut {NoStop}%
\bibitem [{\citenamefont {Childs}\ \emph {et~al.}(2022)\citenamefont {Childs},
  \citenamefont {Li}, \citenamefont {Liu}, \citenamefont {Wang},\ and\
  \citenamefont {Zhang}}]{childs2022quantum}%
  \BibitemOpen
  \bibfield  {author} {\bibinfo {author} {\bibfnamefont {A.~M.}\ \bibnamefont
  {Childs}}, \bibinfo {author} {\bibfnamefont {T.}~\bibnamefont {Li}}, \bibinfo
  {author} {\bibfnamefont {J.-P.}\ \bibnamefont {Liu}}, \bibinfo {author}
  {\bibfnamefont {C.}~\bibnamefont {Wang}},\ and\ \bibinfo {author}
  {\bibfnamefont {R.}~\bibnamefont {Zhang}},\ }\bibfield  {title} {\bibinfo
  {title} {Quantum algorithms for sampling log-concave distributions and
  estimating normalizing constants},\ }\href@noop {} {\bibfield  {journal}
  {\bibinfo  {journal} {arXiv preprint arXiv:2210.06539}\ } (\bibinfo {year}
  {2022})}\BibitemShut {NoStop}%
\bibitem [{\citenamefont {Nielsen}\ and\ \citenamefont
  {Chuang}(2002)}]{nielsen2002quantum}%
  \BibitemOpen
  \bibfield  {author} {\bibinfo {author} {\bibfnamefont {M.~A.}\ \bibnamefont
  {Nielsen}}\ and\ \bibinfo {author} {\bibfnamefont {I.}~\bibnamefont
  {Chuang}},\ }\href@noop {} {\bibinfo {title} {Quantum computation and quantum
  information}} (\bibinfo {year} {2002})\BibitemShut {NoStop}%
\bibitem [{\citenamefont {Babbush}\ \emph {et~al.}(2018)\citenamefont
  {Babbush}, \citenamefont {Gidney}, \citenamefont {Berry}, \citenamefont
  {Wiebe}, \citenamefont {McClean}, \citenamefont {Paler}, \citenamefont
  {Fowler},\ and\ \citenamefont {Neven}}]{babbush2018encoding}%
  \BibitemOpen
  \bibfield  {author} {\bibinfo {author} {\bibfnamefont {R.}~\bibnamefont
  {Babbush}}, \bibinfo {author} {\bibfnamefont {C.}~\bibnamefont {Gidney}},
  \bibinfo {author} {\bibfnamefont {D.~W.}\ \bibnamefont {Berry}}, \bibinfo
  {author} {\bibfnamefont {N.}~\bibnamefont {Wiebe}}, \bibinfo {author}
  {\bibfnamefont {J.}~\bibnamefont {McClean}}, \bibinfo {author} {\bibfnamefont
  {A.}~\bibnamefont {Paler}}, \bibinfo {author} {\bibfnamefont
  {A.}~\bibnamefont {Fowler}},\ and\ \bibinfo {author} {\bibfnamefont
  {H.}~\bibnamefont {Neven}},\ }\bibfield  {title} {\bibinfo {title} {Encoding
  electronic spectra in quantum circuits with linear {T} complexity},\
  }\href@noop {} {\bibfield  {journal} {\bibinfo  {journal} {Physical Review
  X}\ }\textbf {\bibinfo {volume} {8}},\ \bibinfo {pages} {041015} (\bibinfo
  {year} {2018})}\BibitemShut {NoStop}%
\bibitem [{\citenamefont {Arunachalam}\ \emph {et~al.}(2015)\citenamefont
  {Arunachalam}, \citenamefont {Gheorghiu}, \citenamefont {Jochym-O’Connor},
  \citenamefont {Mosca},\ and\ \citenamefont
  {Srinivasan}}]{arunachalam2015robustness}%
  \BibitemOpen
  \bibfield  {author} {\bibinfo {author} {\bibfnamefont {S.}~\bibnamefont
  {Arunachalam}}, \bibinfo {author} {\bibfnamefont {V.}~\bibnamefont
  {Gheorghiu}}, \bibinfo {author} {\bibfnamefont {T.}~\bibnamefont
  {Jochym-O’Connor}}, \bibinfo {author} {\bibfnamefont {M.}~\bibnamefont
  {Mosca}},\ and\ \bibinfo {author} {\bibfnamefont {P.~V.}\ \bibnamefont
  {Srinivasan}},\ }\bibfield  {title} {\bibinfo {title} {On the robustness of
  bucket brigade quantum {RAM}},\ }\href@noop {} {\bibfield  {journal}
  {\bibinfo  {journal} {New Journal of Physics}\ }\textbf {\bibinfo {volume}
  {17}},\ \bibinfo {pages} {123010} (\bibinfo {year} {2015})}\BibitemShut
  {NoStop}%
\bibitem [{\citenamefont {Di~Matteo}\ \emph {et~al.}(2020)\citenamefont
  {Di~Matteo}, \citenamefont {Gheorghiu},\ and\ \citenamefont
  {Mosca}}]{di2020fault}%
  \BibitemOpen
  \bibfield  {author} {\bibinfo {author} {\bibfnamefont {O.}~\bibnamefont
  {Di~Matteo}}, \bibinfo {author} {\bibfnamefont {V.}~\bibnamefont
  {Gheorghiu}},\ and\ \bibinfo {author} {\bibfnamefont {M.}~\bibnamefont
  {Mosca}},\ }\bibfield  {title} {\bibinfo {title} {Fault-tolerant resource
  estimation of quantum random-access memories},\ }\href@noop {} {\bibfield
  {journal} {\bibinfo  {journal} {IEEE Transactions on Quantum Engineering}\
  }\textbf {\bibinfo {volume} {1}},\ \bibinfo {pages} {1} (\bibinfo {year}
  {2020})}\BibitemShut {NoStop}%
\bibitem [{\citenamefont {Mohseni}\ \emph {et~al.}(2024)\citenamefont
  {Mohseni}, \citenamefont {Scherer}, \citenamefont {Johnson}, \citenamefont
  {Wertheim}, \citenamefont {Otten}, \citenamefont {Aadit}, \citenamefont
  {Bresniker}, \citenamefont {Camsari}, \citenamefont {Chapman}, \citenamefont
  {Chatterjee} \emph {et~al.}}]{mohseni2024build}%
  \BibitemOpen
  \bibfield  {author} {\bibinfo {author} {\bibfnamefont {M.}~\bibnamefont
  {Mohseni}}, \bibinfo {author} {\bibfnamefont {A.}~\bibnamefont {Scherer}},
  \bibinfo {author} {\bibfnamefont {K.~G.}\ \bibnamefont {Johnson}}, \bibinfo
  {author} {\bibfnamefont {O.}~\bibnamefont {Wertheim}}, \bibinfo {author}
  {\bibfnamefont {M.}~\bibnamefont {Otten}}, \bibinfo {author} {\bibfnamefont
  {N.~A.}\ \bibnamefont {Aadit}}, \bibinfo {author} {\bibfnamefont {K.~M.}\
  \bibnamefont {Bresniker}}, \bibinfo {author} {\bibfnamefont {K.~Y.}\
  \bibnamefont {Camsari}}, \bibinfo {author} {\bibfnamefont {B.}~\bibnamefont
  {Chapman}}, \bibinfo {author} {\bibfnamefont {S.}~\bibnamefont {Chatterjee}},
  \emph {et~al.},\ }\bibfield  {title} {\bibinfo {title} {How to build a
  quantum supercomputer: Scaling challenges and opportunities},\ }\href@noop {}
  {\bibfield  {journal} {\bibinfo  {journal} {arXiv preprint arXiv:2411.10406}\
  } (\bibinfo {year} {2024})}\BibitemShut {NoStop}%
\bibitem [{\citenamefont {Pavliotis}(2014)}]{pavliotis2014stochastic}%
  \BibitemOpen
  \bibfield  {author} {\bibinfo {author} {\bibfnamefont {G.~A.}\ \bibnamefont
  {Pavliotis}},\ }\href@noop {} {\emph {\bibinfo {title} {Stochastic processes
  and applications: diffusion processes, the {F}okker-{P}lanck and Langevin
  equations}}},\ Vol.~\bibinfo {volume} {60}\ (\bibinfo  {publisher}
  {Springer},\ \bibinfo {address} {London, UK},\ \bibinfo {year}
  {2014})\BibitemShut {NoStop}%
\bibitem [{\citenamefont {Krizhevsky}(2009)}]{Krizhevsky09learningmultiple}%
  \BibitemOpen
  \bibfield  {author} {\bibinfo {author} {\bibfnamefont {A.}~\bibnamefont
  {Krizhevsky}},\ }\href@noop {} {\emph {\bibinfo {title} {Learning multiple
  layers of features from tiny images}}},\ \bibinfo {type} {Tech. Rep.}\
  (\bibinfo {year} {2009})\BibitemShut {NoStop}%
\bibitem [{\citenamefont {Tieleman}\ and\ \citenamefont
  {Hinton}(2009)}]{tieleman2009using}%
  \BibitemOpen
  \bibfield  {author} {\bibinfo {author} {\bibfnamefont {T.}~\bibnamefont
  {Tieleman}}\ and\ \bibinfo {author} {\bibfnamefont {G.}~\bibnamefont
  {Hinton}},\ }\bibfield  {title} {\bibinfo {title} {Using fast weights to
  improve persistent contrastive divergence},\ }in\ \href@noop {} {\emph
  {\bibinfo {booktitle} {Proceedings of the 26th annual international
  conference on machine learning}}}\ (\bibinfo {year} {2009})\ pp.\ \bibinfo
  {pages} {1033--1040}\BibitemShut {NoStop}%
\bibitem [{\citenamefont {Kingma}\ and\ \citenamefont
  {Ba}(2014)}]{kingma2014method}%
  \BibitemOpen
  \bibfield  {author} {\bibinfo {author} {\bibfnamefont {D.~P.}\ \bibnamefont
  {Kingma}}\ and\ \bibinfo {author} {\bibfnamefont {J.}~\bibnamefont {Ba}},\
  }\bibfield  {title} {\bibinfo {title} {Adam: A method for stochastic
  optimization},\ }\href@noop {} {\  (\bibinfo {year} {2014})},\ \bibinfo
  {note} {cite arxiv:1412.6980 Comment: Published as a conference paper at the
  3rd International Conference for Learning Representations, San Diego,
  2015}\BibitemShut {NoStop}%
\bibitem [{\citenamefont {Madry}\ \emph {et~al.}(2017)\citenamefont {Madry},
  \citenamefont {Makelov}, \citenamefont {Schmidt}, \citenamefont {Tsipras},\
  and\ \citenamefont {Vladu}}]{madry2017towards}%
  \BibitemOpen
  \bibfield  {author} {\bibinfo {author} {\bibfnamefont {A.}~\bibnamefont
  {Madry}}, \bibinfo {author} {\bibfnamefont {A.}~\bibnamefont {Makelov}},
  \bibinfo {author} {\bibfnamefont {L.}~\bibnamefont {Schmidt}}, \bibinfo
  {author} {\bibfnamefont {D.}~\bibnamefont {Tsipras}},\ and\ \bibinfo {author}
  {\bibfnamefont {A.}~\bibnamefont {Vladu}},\ }\bibfield  {title} {\bibinfo
  {title} {Towards deep learning models resistant to adversarial attacks},\
  }\href@noop {} {\bibfield  {journal} {\bibinfo  {journal} {arXiv preprint
  arXiv:1706.06083}\ } (\bibinfo {year} {2017})}\BibitemShut {NoStop}%
\bibitem [{\citenamefont {Athalye}\ \emph {et~al.}(2018)\citenamefont
  {Athalye}, \citenamefont {Carlini},\ and\ \citenamefont
  {Wagner}}]{athalye2018obfuscated}%
  \BibitemOpen
  \bibfield  {author} {\bibinfo {author} {\bibfnamefont {A.}~\bibnamefont
  {Athalye}}, \bibinfo {author} {\bibfnamefont {N.}~\bibnamefont {Carlini}},\
  and\ \bibinfo {author} {\bibfnamefont {D.}~\bibnamefont {Wagner}},\
  }\bibfield  {title} {\bibinfo {title} {Obfuscated gradients give a false
  sense of security: Circumventing defenses to adversarial examples},\ }in\
  \href@noop {} {\emph {\bibinfo {booktitle} {International conference on
  machine learning}}}\ (\bibinfo {organization} {PMLR},\ \bibinfo {year}
  {2018})\ pp.\ \bibinfo {pages} {274--283}\BibitemShut {NoStop}%
\bibitem [{\citenamefont {DARPA}(2021)}]{darpa2021benchmark}%
  \BibitemOpen
  \bibfield  {author} {\bibinfo {author} {\bibnamefont {DARPA}},\ }\href@noop
  {} {\bibinfo {title} {Quantifying utility of quantum computers}},\ \bibinfo
  {howpublished} {Retrieved April 16, 2023, from
  \url{https://www.darpa.mil/news-events/2021-04-02}} (\bibinfo {year}
  {2021}),\ \bibinfo {note} {accessed on April 16, 2023}\BibitemShut {NoStop}%
\bibitem [{\citenamefont {Ho}\ \emph {et~al.}(2020)\citenamefont {Ho},
  \citenamefont {Jain},\ and\ \citenamefont {Abbeel}}]{ho2020denoising}%
  \BibitemOpen
  \bibfield  {author} {\bibinfo {author} {\bibfnamefont {J.}~\bibnamefont
  {Ho}}, \bibinfo {author} {\bibfnamefont {A.}~\bibnamefont {Jain}},\ and\
  \bibinfo {author} {\bibfnamefont {P.}~\bibnamefont {Abbeel}},\ }\bibfield
  {title} {\bibinfo {title} {Denoising diffusion probabilistic models},\
  }\href@noop {} {\bibfield  {journal} {\bibinfo  {journal} {Advances in Neural
  Information Processing Systems}\ }\textbf {\bibinfo {volume} {33}},\ \bibinfo
  {pages} {6840} (\bibinfo {year} {2020})}\BibitemShut {NoStop}%
\bibitem [{\citenamefont {Desislavov}\ \emph {et~al.}(2021)\citenamefont
  {Desislavov}, \citenamefont {Mart{\'\i}nez-Plumed},\ and\ \citenamefont
  {Hern{\'a}ndez-Orallo}}]{desislavov2021compute}%
  \BibitemOpen
  \bibfield  {author} {\bibinfo {author} {\bibfnamefont {R.}~\bibnamefont
  {Desislavov}}, \bibinfo {author} {\bibfnamefont {F.}~\bibnamefont
  {Mart{\'\i}nez-Plumed}},\ and\ \bibinfo {author} {\bibfnamefont
  {J.}~\bibnamefont {Hern{\'a}ndez-Orallo}},\ }\bibfield  {title} {\bibinfo
  {title} {Compute and energy consumption trends in deep learning inference},\
  }\href@noop {} {\bibfield  {journal} {\bibinfo  {journal} {arXiv preprint
  arXiv:2109.05472}\ } (\bibinfo {year} {2021})}\BibitemShut {NoStop}%
\bibitem [{\citenamefont {Li}\ \emph {et~al.}(2023)\citenamefont {Li},
  \citenamefont {Yang}, \citenamefont {Islam},\ and\ \citenamefont
  {Ren}}]{li2023making}%
  \BibitemOpen
  \bibfield  {author} {\bibinfo {author} {\bibfnamefont {P.}~\bibnamefont
  {Li}}, \bibinfo {author} {\bibfnamefont {J.}~\bibnamefont {Yang}}, \bibinfo
  {author} {\bibfnamefont {M.~A.}\ \bibnamefont {Islam}},\ and\ \bibinfo
  {author} {\bibfnamefont {S.}~\bibnamefont {Ren}},\ }\bibfield  {title}
  {\bibinfo {title} {Making {AI} less thirsty: Uncovering and addressing the
  secret water footprint of {AI} models},\ }\href@noop {} {\bibfield  {journal}
  {\bibinfo  {journal} {arXiv preprint arXiv:2304.03271}\ } (\bibinfo {year}
  {2023})}\BibitemShut {NoStop}%
\end{thebibliography}%

\end{document}